\documentclass[10pt,twocolumn,letterpaper]{article}
\usepackage{cvpr}   

\usepackage{amsmath,amsfonts,bm,mathtools}
\usepackage{graphicx}
\usepackage{booktabs}
\usepackage{multirow,makecell}
\usepackage{tabularx}
\usepackage{colortbl}
\usepackage{algorithm}
\usepackage{algpseudocode}
\usepackage{subcaption}
\usepackage{siunitx}
\definecolor{cvprblue}{rgb}{0.21,0.49,0.74}
\usepackage[pagebackref,breaklinks,colorlinks,allcolors=cvprblue]{hyperref}

\usepackage[dvipsnames]{xcolor}
\definecolor{mygray}{HTML}{E9F1F6}

\usepackage{pifont}
\usepackage{amsmath,amsfonts,bm}

\def\eqref#1{equation~\ref{#1}}
\def\1{\bm{1}}

\def\rvc{{\mathbf{c}}}

\def\rve{{\mathbf{e}}}

\def\rvt{{\mathbf{t}}}

\def\rvv{{\mathbf{v}}}
\def\rvw{{\mathbf{w}}}

\def\rvz{{\mathbf{z}}}

\def\rmA{{\mathbf{A}}}

\def\rmF{{\mathbf{F}}}

\def\rmI{{\mathbf{I}}}

\def\rmM{{\mathbf{M}}}

\def\rmP{{\mathbf{P}}}

\def\rmT{{\mathbf{T}}}

\DeclareMathAlphabet{\mathsfit}{\encodingdefault}{\sfdefault}{m}{sl}
\SetMathAlphabet{\mathsfit}{bold}{\encodingdefault}{\sfdefault}{bx}{n}
\def\gE{{\mathcal{E}}}

\def\gG{{\mathcal{G}}}

\def\gL{{\mathcal{L}}}

\def\gP{{\mathcal{P}}}

\def\gT{{\mathcal{T}}}

\def\gV{{\mathcal{V}}}

\title{Unifying Adaptive Generative Augmentation and Quality-Driven Supervision for Contrastive Representation Learning}

\author{
Xiaojie Li$^{1,2}$ \quad
Bei Wang$^{1}$ \quad
Wei Liu$^{1}$ \quad
Jianlong Wu$^{1}$ \quad
Yue Yu$^{2}$ \quad
Liqiang Nie$^{1}$ \quad
Min Zhang$^{1}$ \\
{\normalsize
$^1$Harbin Institute of Technology, Shenzhen \quad
$^2$Peng Cheng Laboratory
} \\
{\tt\small
xiaojieli0903@gmail.com,
wangbei010118@gmail.com,
liuwei030224@gmail.com,
} \\
{\tt\small
wujianlong@hit.edu.cn
yuy@pcl.ac.cn,
nieliqiang@gmail.com,
zhangmin2021@hit.edu.cn
}
}

\begin{document}

\maketitle
\begin{abstract}
The efficacy of contrastive representation learning fundamentally hinges on the quality of positive pairs, which must strike a delicate balance between semantic fidelity and distributional diversity. However, current methods face critical limitations: on the construction side, handcrafted augmentations offer limited variation, while rigid generative approaches often induce semantic drift due to a lack of fine-grained controllability; on the learning side, the standard “augment-and-trust” paradigm treats all pairs equally, leading to suboptimal supervision prone to overfitting on noisy samples. To tackle these challenges, we propose GenView++, a unified framework that integrates controllable generative synthesis with quality-aware supervision. First, to construct superior pairs, we introduce a Multi-Source Adaptive View Generation mechanism. Leveraging the latent conditioning of pre-trained diffusion models, we modulate the noise injection levels and text guidance scales based on input intrinsic properties—specifically, visual saliency derived from self-supervised feature priors and caption complexity. This ensures the synthesis of diverse yet semantically coherent views. Second, to optimize pair utilization, we propose a Quality-Driven Contrastive Learning mechanism. By explicitly quantifying cross-modal semantic alignment and visual novelty, we implement a soft weighting scheme that prioritizes informative pairs while suppressing redundant or misaligned ones. Extensive experiments demonstrate the robustness of GenView++ across both vision and vision–language tasks. For vision representation learning, it improves MoCov2 by +2.5\% on ImageNet linear classification. For vision–language learning, it raises the average zero-shot classification accuracy by +12.31\% over CLIP and +5.31\% over SLIP across ten datasets, and further improves Flickr30k text retrieval R@5 by +3.2\%. The code is available at \url{https://github.com/xiaojieli0903/GenViewPlusPlus}.
\end{abstract}

\begin{figure*}
 \centering
 \includegraphics[width=\linewidth]{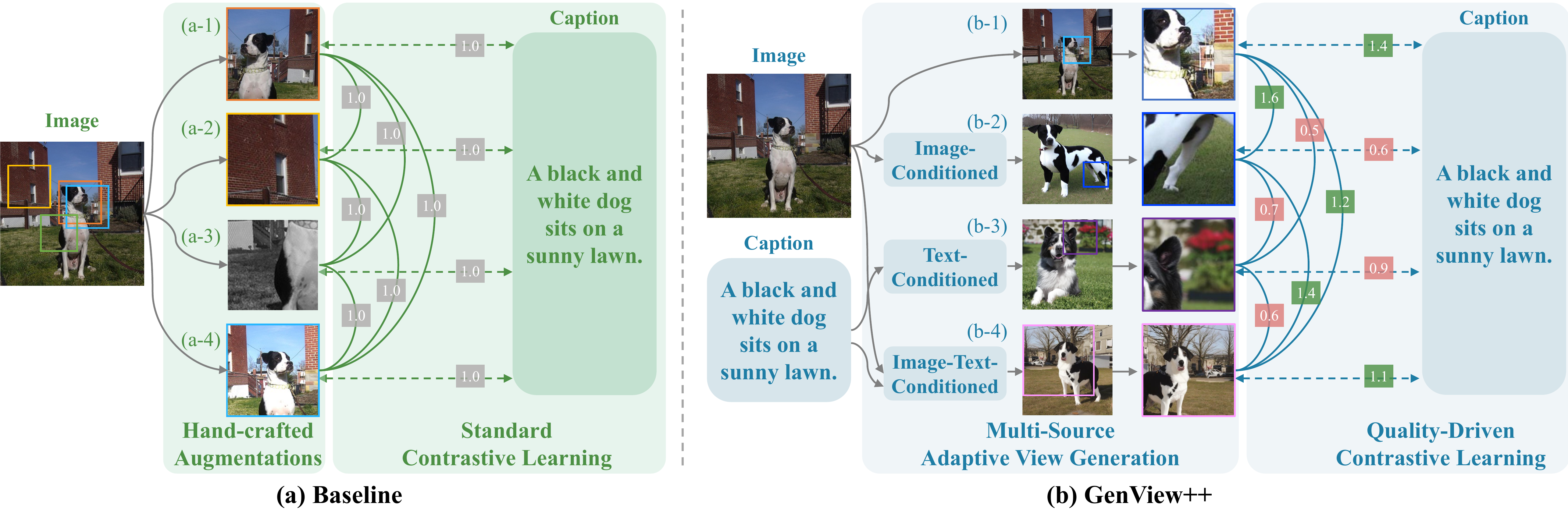}
 \vspace{-6mm}
\caption{\textbf{Motivation.} \textbf{(a)} Standard contrastive learning methods rely on handcrafted augmentations, which often yield limited diversity or risk semantic distortions, and lack pair-level quality control during training. \textbf{(b)} \textbf{GenView++} addresses these limitations via a unified framework: (i) \textbf{Multi-Source Adaptive View Generation} leverages pretrained generative models to synthesize diverse, semantically consistent views across modalities, and (ii) \textbf{Quality-Driven Contrastive Learning} dynamically reweighting image-image and image-text pairs based on their semantic alignment and visual diversity.}
 \label{fig:motivation_GenView++}
  \vspace{-4mm}
\end{figure*}

\section{Introduction}
\label{sec:intro}
Self-supervised learning (SSL) has emerged as a fundamental paradigm for learning robust and generalizable representations from large-scale unlabeled data. Specifically, contrastive learning (CL)~\cite{chen2020simple, caron2020unsupervised} has achieved remarkable success in both \textit{vision}~\cite{he2020momentum, grill2020bootstrap} and \textit{vision-language}~\cite{radford2021learning, li2022blip} domains. The core principle of CL is to learn an invariant embedding space by maximizing the similarity between different views of the same instance (positive pairs) while pushing them apart from others. Consequently, the efficacy of CL fundamentally hinges on the quality of positive pairs, which must strike a delicate trade-off: they must be \textit{semantically consistent} to ensure true invariance, yet \textit{sufficiently diverse} to foster robust generalization.

However, constructing such optimal pairs remains a persistent challenge constrained by two factors.
\textbf{First, on the construction side}, traditional methods rely on handcrafted augmentations (e.g., cropping, jittering). These transformations modify only surface-level statistics, often yielding redundant views with \textit{limited diversity} (Fig.~\ref{fig:motivation_GenView++} a-1) or risking \textit{semantic corruption} when applied aggressively (a-2). While recent generative approaches~\cite{tian2023stablerep, zhang2022expanding} leverage pretrained diffusion models~\cite{rombach2022high} to synthesize diverse views, they typically lack fine-grained \textit{controllability}. Relying on fixed noise levels or unconditional guidance often leads to \textbf{semantic drift}—where generated views deviate significantly from the original semantics—or redundant outputs that offer little learning signal. Furthermore, most existing methods are constrained by a single conditioning source~\cite{zhang2022expanding, ye2023exploiting, tian2023stablerep}, utilizing either the image or the text, but not both. This unimodal approach overlooks the synergistic information inherent in joint image-text pairs, missing the opportunity to synthesize views that are simultaneously consistent with the visual content and aligned with textual semantics.
\textbf{Second, on the learning side}, current pipelines operate on a uniform ``augment-and-trust" assumption, treating every positive pair as equally important regardless of its quality (Fig.~\ref{fig:motivation_GenView++} a). This indiscriminate supervision forces models to overfit to false positives generated by aggressive augmentations or noisy web data~\cite{schuhmann2022laion}, ultimately degrading the discriminative power of the learned features.

To address the challenges of construction diversity and supervision quality in vision-only pretraining, our preliminary work, \textbf{GenView}~\cite{li2024genview} (ECCV 2024), pioneered a controllable framework using saliency-guided adaptive noise injection. While effective for visual tasks, GenView is fundamentally constrained by its \textit{unimodal nature}. It neglects the rich semantic context embedded in text, thereby limiting its efficacy for broader vision-language representation learning. Furthermore, relying solely on image-based metrics renders the framework blind to \textit{cross-modal semantic drifts}—cases where a generated image maintains visual coherence but diverges from its textual description.

To bridge this gap and generalize the principle of adaptive control to the multimodal domain, we propose \textbf{GenView++}. Unlike its predecessor, GenView++ is a \textbf{unified framework} that harmonizes multi-source generative augmentation with cross-modal quality supervision. It establishes a robust pipeline for both vision and vision-language representation learning by introducing two innovations:

\textbf{First}, we introduce a \textbf{Multi-Source Adaptive View Generation} module. Unlike rigid synthesis methods that apply fixed parameters, this offline mechanism dynamically modulates the generative process based on the input's intrinsic properties to balance semantic fidelity with diversity. It integrates three adaptively controlled strategies leveraging the latent conditioning of unCLIP models:
\textbf{(1) Image-Conditioned Mode} (Fig.~\ref{fig:motivation_GenView++} b-2) enhances visual invariance via a \textit{saliency-aware noise injection} strategy. Building on visual saliency priors (derived from self-supervised features), we assign weaker perturbations to low-saliency subjects to preserve identity, while allowing stronger perturbations for high-saliency subjects to enrich background diversity.
\textbf{(2) Text-Conditioned Mode} (Fig.~\ref{fig:motivation_GenView++} b-3) exploits textual semantics, adaptively tuning the classifier-free guidance scale based on \textit{caption complexity}. Intuitively, simple or abstract captions trigger stronger guidance to prevent semantic drift, whereas complex, descriptive captions allow for relaxed guidance to foster creative transformations without losing context.
\textbf{(3) Image–Text-Conditioned Mode} (Fig.~\ref{fig:motivation_GenView++} b-4) bridges the modality gap by synergistically adapting both image noise and text guidance. This joint control produces views that are visually diverse, semantically consistent with the image, and textually grounded in the caption.

\textbf{Second}, we introduce a \textbf{Quality-Driven Contrastive Learning} mechanism to bridge the gap between pair construction and utilization. Unlike the conventional ``augment-and-trust'' paradigm that treats all pairs equally, our approach explicitly evaluates and reweights pairs based on two complementary dimensions:
(1) \textbf{Semantic Fidelity}, measured by foreground consistency (for image pairs) or CLIP-based cross-modal alignment (for image-text pairs), ensures that the generated view does not suffer from semantic drift.
(2) \textbf{Informative Diversity}, quantified by background novelty or visual variation, ensures that the view contributes non-trivial gradients rather than mere redundancy.
By combining these metrics, we dynamically modulate the training objective: high-quality pairs (high fidelity and high diversity) are amplified (green scores in Fig.~\ref{fig:motivation_GenView++} b), while redundant or misaligned false positives are suppressed. This mechanism effectively prevents the model from overfitting to generative noise, ensuring robust representation learning.

In summary, this manuscript represents a substantial extension of our conference version~\cite{li2024genview}. While GenView focused exclusively on image-conditioned augmentation, GenView++ generalizes the methodology into a comprehensive multimodal framework. Specifically, we contribute: (1) novel \textbf{text-conditioned} and \textbf{image-text-conditioned} adaptive generation strategies to bridge the modality gap; (2) a new \textbf{cross-modal quality assessment metric} to regulate vision-language alignment; and (3) significantly expanded evaluations on \textbf{vision-language benchmarks} including zero-shot classification and retrieval. This version provides a more systemic solution to the noise-overfitting problem in contrastive learning.

Our contributions are as follows:
\begin{itemize}
\item We propose GenView++, a unified framework that harmonizes adaptive generative augmentation with quality-driven supervision to enhance both vision and vision-language representation learning.
\item We introduce a multi-source adaptive view generation module. By dynamically modulating latent noise and guidance parameters based on input characteristics, it effectively balances the trade-off between semantic fidelity and diversity.
\item We develop a quality-driven contrastive mechanism that reweights training pairs based on cross-modal alignment and visual diversity, preventing overfitting to noisy samples.
\item Extensive experiments demonstrate the effectiveness of GenView++. For vision tasks, it improves MoCov2 by \textbf{+2.5\%} on ImageNet; for vision-language tasks, it outperforms CLIP and StableRep, raising zero-shot classification accuracy by \textbf{+12.31\%} over CLIP.
\end{itemize}
\section{Related Work}
\label{sec:related}

\subsection{Contrastive Representation Learning}
Self-supervised learning (SSL) has evolved from heuristic pretext tasks~\cite{noroozi2016unsupervised, gidaris2018unsupervised, vincent2008extracting, pathak2016context,lim2023scl,fonseca2022robust,biscione2022learning,zhang2025mmc} to two dominant paradigms: Masked Image Modeling (MIM) and Contrastive Learning (CL). While MIM approaches like MAE~\cite{he2022masked} excel in fine-tuning tasks by reconstructing pixel-level details, CL~\cite{chen2020simple, he2020momentum, wu2018unsupervised, oord2018representation, tian2019contrastive, chen2020improved, bardes2022vicregl, assran2023self} generally yields superior linearly separable global representations and robust zero-shot transferability. By maximizing the similarity between positive pairs while pushing apart negatives, CL learns invariant features.
Non-contrastive approaches~\cite{caron2020unsupervised, caron2018deep, asano2020self, li2020prototypical, chen2020exploring, grill2020bootstrap, zbontar2021barlow, ermolov2021whitening} further obviate the need for negative pairs via asymmetric networks or redundancy reduction. In the vision-language (VL) domain, foundational models such as CLIP~\cite{radford2021learning}, ALIGN~\cite{jia2021scaling}, and BLIP~\cite{li2022blip} extend this principle to cross-modal pairs from massive web datasets.
Despite their success, these methods remain constrained by \textit{data quality}. On the \textit{construction side}, reliance on handcrafted augmentations offers limited diversity and risks semantic distortion~\cite{shorten2019survey, wei2019eda}. On the \textit{learning side}, standard pipelines typically treat all training pairs as equally reliable. This uniform supervision is suboptimal for noisy web-crawled datasets (e.g., CC12M~\cite{changpinyo2021conceptual}, LAION~\cite{schuhmann2022laion}), where weak image-text alignment can lead to representational degradation.

\subsection{Positive Pair Construction}
The efficacy of CL hinges on the quality of positive pairs. Existing approaches generally fall into two categories: recomposition-based and generation-based.

\vspace{1mm}\noindent \textbf{Recomposition-based Augmentation.}
Early works focused on task-aware or saliency-preserving augmentations~\cite{tian2020makes, selvaraju2021casting} to avoid cropping out objects. More recent approaches utilize the dataset statistics to construct views. SwAV~\cite{caron2020unsupervised} introduces a multi-crop strategy, combining global and local crops to encourage multi-scale consistency. VICRegL~\cite{bardes2022vicregl} and LoGo~\cite{zhang2022leverage} further explicitly align global representations with local features.
A distinct paradigm, represented by NNCLR~\cite{dwibedi2021little}, constructs positive pairs by retrieving nearest neighbors from the dataset rather than augmenting the instance itself. In the VL domain, strategies include semantic-preserving perturbations~\cite{tang2020semantic} and multimodal mixing methods like MDA~\cite{xu2020mda} and MixGen~\cite{hao2023mixgen}.
However, whether using multi-crop or nearest neighbors, these methods are fundamentally bounded by the \textit{original dataset manifold}. They effectively \textit{interpolate} within seen data but cannot create truly novel visual concepts (e.g., unseen poses, lighting, or backgrounds). Furthermore, retrieval-based methods risk introducing false positives if the dataset is sparse or class-imbalanced.

\vspace{1mm}\noindent \textbf{Generation-based Augmentation.}
Generative models offer a paradigm shift by synthesizing entirely new views, effectively \textit{extrapolating} the data manifold. Early approaches trained generators (e.g., GANs) from scratch on the target dataset~\cite{tamkin2020viewmaker, astolfi2023instance, yang2022local, kim2023neural, zang2023boosting}, which remains self-limiting.
Recent works leverage large-scale pretrained diffusion models~\cite{rombach2022high}, injecting external visual priors to synthesize photorealistic views with diverse semantics~\cite{tian2023stablerep, he2022synthetic, shipard2023diversity, trabucco2023effective, jahanian2021generative, zhang2023free, wang2024enhance, patel2024tripletclip, fu2024dreamda}. For instance, StableRep~\cite{tian2023stablerep} learns invariances by contrasting multiple synthetic images generated from text.
Despite their promise, existing generative approaches lack \textit{fine-grained controllability}. They typically rely on fixed noise levels~\cite{tian2023stablerep, he2022synthetic} or random sampling~\cite{shipard2023diversity, trabucco2023effective}, which often leads to \textbf{semantic drift} (when noise is too high) or redundancy (when noise is too low). GenView++ addresses this by introducing a \textit{multi-source adaptive generation} module. We dynamically modulate noise injection and guidance scales based on input saliency and complexity, ensuring synthesized views are both diverse and semantically faithful.

\subsection{Dynamic Sample Selection and Reweighting}
Compounding the construction challenges is the indiscriminate utilization of pairs. Standard pipelines operate on an ``augment-and-trust'' principle, treating every pair as equally valuable.
To address this, extensive research has explored dynamic sample selection. In vision SSL, methods like synthetic hard negatives~\cite{dong2024synthetic} and hard negative mixing~\cite{kalantidis2020hard} focus on mining informative \textit{negative} samples. Others propose mining potential positives~\cite{dong2024rethinking} or reweighting loss to control false negatives~\cite{wang2023weighted}. In the VL domain, hardness-weighted mechanisms~\cite{jiang2023vision, lan2025llave} regulate cross-modal similarity to focus on hard-to-distinguish pairs.
While these methods primarily focus on \textit{mining hardness} or \textit{filtering noise} within static datasets, GenView++ tackles a distinct challenge specific to generative augmentation: the \textbf{fidelity-diversity trade-off}. Instead of simply prioritizing hard samples (which might be semantic outliers in generative contexts), our mechanism functions as a \textbf{soft denoising curriculum}. It explicitly evaluates synthesized views, rewarding pairs that exhibit both high semantic alignment and sufficient visual diversity, thereby preventing overfitting to generative artifacts.
\begin{figure*}
 \centering
 \includegraphics[width=1\linewidth]{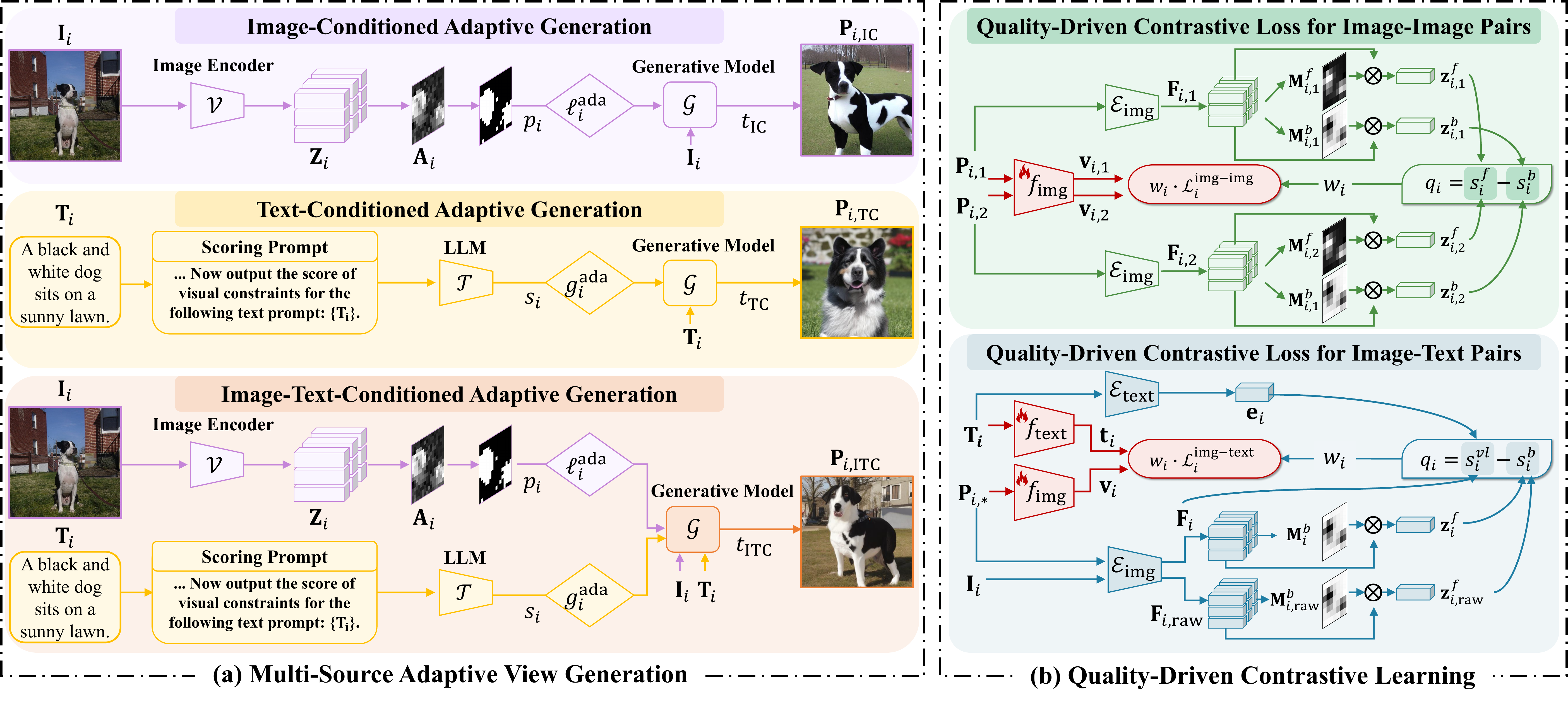}
 \vspace{-6mm}
\caption{\textbf{Overview of the GenView++ Framework.} The framework enhances contrastive learning via two synergistic modules:
\textbf{(a) Offline Multi-Source Adaptive View Generation.} Diverse views are synthesized using three adaptive strategies: \textbf{Image-Conditioned} generation perturbs visual embeddings of the source image $\rmI_i$ using adaptive noise $\ell^\text{ada}_i$ based on foreground ratio $p_i$ to produce $\rmP_{i,\text{IC}}$; \textbf{Text-Conditioned} generation adjusts guidance scale $g^\text{ada}_i$ based on caption complexity $s_i$ for $\rmP_{i,\text{TC}}$; \textbf{Image-Text-Conditioned} generation jointly leverages both modalities for $\rmP_{i,\text{ITC}}$.
\textbf{(b) Online Quality-Driven Contrastive Learning.} A dynamic reweighting scheme adjusts contrastive losses based on quality scores. For intra-modal loss $\gL_i^{\text{II}}$, $q_i$ is derived from foreground consistency ($s_i^f$) and background diversity ($-s_i^b$).
For cross-modal loss $\gL_i^{\text{IT}}$, $q_i$ reflects cross-modal alignment ($s_i^\text{IT}$) and visual diversity ($-s_i^b$). The quality scores are normalized as reweighting factors $w_i$ to prioritize high-quality pairs and suppress noisy pairs.}
\label{fig:framework_GenView++}
\end{figure*}

\section{Preliminaries}
\label{method:preliminaries}
\subsection{Contrastive Representation Learning}
\label{pre:vision_pretraining}
Contrastive learning aims to learn an invariant embedding space where semantically similar samples are pulled together while dissimilar ones are pushed apart. 
In the vision-only domain, given a batch of $N$ images, standard methods construct positive pairs via data augmentation. For each image $\rmI_i$, two augmented views $\rmP_{i,1} = t_1(\rmI_i)$ and $\rmP_{i,2} = t_2(\rmI_i)$ are generated via random transformations $t_1, t_2 \sim \mathcal{T}$.
These views are processed by an encoder $\gE_\text{img}(\cdot)$ and a projection head to obtain feature vectors $\rvv_{i,1}, \rvv_{i,2}$. 
Most methods (e.g., MoCo~\cite{he2020momentum}) optimize these features using the Noise Contrastive Estimation (NCE) loss:
\begin{equation}\label{eq:loss_ssl_nce}
\gL_{i}^\text{II} = -\operatorname{log} \frac{\operatorname{exp}~(\rvv_{i,1} \cdot \rvv_{i,2}/ \tau) }{\sum_{k=1}^N \operatorname{exp}~(\rvv_{i,1} \cdot \rvv_{k} / \tau ) },
\end{equation}
where $\tau$ is a temperature parameter and $\rvv_k$ represents features from other samples in the batch (serving as negatives). Our framework is objective-agnostic and compatible with non-contrastive losses such as the cosine similarity in BYOL~\cite{grill2020bootstrap} or KL-divergence in SwAV~\cite{caron2020unsupervised}.

\subsection{Vision-Language Representation Learning}
\label{pre:vision_language_pretraining}
Vision–language pretraining extends this principle to align visual and textual modalities. Given a batch of $N$ image-text pairs $\{(\rmI_i, \rmT_i)\}$, an image encoder $\gE_\text{img}$ and a text encoder $\gE_\text{text}$ extract visual features $\rvv_i$ and textual features $\rvt_i$, respectively.
The standard objective maximizes the similarity for matched pairs $(\rvv_i, \rvt_i)$ while minimizing it for mismatched ones using a symmetric cross-entropy loss:
\begin{equation}
\gL_{i}^\text{IT} = \underbrace{- \log \frac{e^{\langle \rvv_i, \rvt_i \rangle /\tau}}{\sum_{j} e^{\langle \rvv_i, \rvt_j \rangle /\tau}}}_{\gL_{i, \text{i2t}}} \underbrace{- \log \frac{e^{\langle \rvt_i, \rvv_i \rangle /\tau}}{\sum_{j} e^{\langle \rvt_i, \rvv_j \rangle /\tau}}}_{\gL_{i, \text{t2i}}},
\label{eq:vl_loss_sym}
\end{equation}
where $\langle \cdot, \cdot \rangle$ denotes cosine similarity. GenView++ enhances these objectives by improving the quality of the input positive pairs and reweighting $\gL^\text{II}$ and $\gL^\text{IT}$.

\section{The GenView++ Framework}
\label{method:framework}

\subsection{Unified Architecture}
\label{method:overall}
GenView++ is designed as a unified framework that bridges the gap between \textit{generative data construction} and \textit{discriminative representation learning}. As illustrated in Fig.~\ref{fig:framework_GenView++}, it consists of two decoupled yet synergistic modules:

\vspace{1mm}\noindent\textbf{(1) Multi-Source Adaptive View Generation (Offline).} 
Instead of relying on fixed augmentation policies, this module leverages the rich priors of pre-trained diffusion models to expand the training data manifold. Crucially, we introduce an \textbf{adaptive control mechanism} that modulates latent generation parameters (noise level and guidance scale) based on instance-specific properties, ensuring that generated views maintain semantic fidelity while maximizing diversity.
\textit{Remark on Efficiency:} This process is performed entirely \textit{offline}. It represents a one-time computational investment to create a static, reusable pool of high-quality views, incurring no additional overhead during the online pretraining phase.

\vspace{1mm}\noindent\textbf{(2) Quality-Driven Contrastive Learning (Online).} 
To mitigate the inevitable noise from generative synthesis and web-crawled data, this module functions as a \textbf{soft denoising curriculum}. It dynamically assesses the quality of each training pair—measuring both semantic alignment and visual informativeness—and reweights the contrastive objective to prioritize clean, informative signals.

The framework supports two pretraining settings:

\vspace{1mm} \noindent \textbf{GenView++ (V) for Vision-Only Pretraining:} Adopts a \textit{Real-to-Synthetic} pairing strategy. Each real image $\rmI_i$ is paired with an adaptively generated view $\rmI_{i,\text{IC}}$ (Image-Conditioned). Unlike methods that utilize purely synthetic pairs, we retain the original real image $\rmI_i$ as a \textbf{distributional anchor}. This prevents feature drift caused by domain gaps between the generative model and the target dataset, ensuring the learned representation remains grounded in the real-world distribution. The augmented pair $\{\rmP_{i,\text{ori}}, \rmP_{i,\text{IC}}\}$ is optimized via the quality-weighted intra-modal loss $\gL^{\text{II,QD}}_i$.
    
\vspace{1mm} \noindent \textbf{GenView++ (VL) for Vision-Language Pretraining:} Constructs a comprehensive semantic view set for each image-text pair $(\rmI_i, \rmT_i)$ by activating all three adaptive generation modes. This results in an expanded positive set $\gP_i$:
    \begin{equation}
    \label{eq:pos_set}
    \gP_i=\{\rmP_{i,\text{ori}}, \underbrace{\rmP_{i,\text{IC}}}_{\text{Image-Cond.}}, \underbrace{\rmP_{i,\text{TC}}}_{\text{Text-Cond.}}, \underbrace{\rmP_{i,\text{ITC}}}_{\text{Joint-Cond.}}\}.
    \end{equation}
Training is guided by joint quality-weighted objectives: intra-modal contrast ($\gL^\text{II, QD}_i$) enforces robustness to visual variations, while cross-modal contrast ($\gL^\text{IT, QD}_i$) aligns the generated views with the caption embedding $\rvt_i$ to ensure semantic consistency.

\subsection{Multi-Source Adaptive View Generation}
\label{method:ada}
Standard handcrafted or generative augmentations often suffer from a rigid trade-off between diversity and fidelity. To resolve this, we propose a general adaptive controllability mechanism that modulates the \textit{conditioning space} of latent diffusion models. Instead of treating generation as a black box, we dynamically adjust the stochasticity of the input conditions based on their intrinsic properties.

\subsubsection{Mechanisms for Controllable Synthesis}
\label{sec:method_ada_parameters}
We instantiate our framework using Stable unCLIP 2.1~\cite{rombach2022high, ramesh2022hierarchical} as the generative backbone $\mathcal{G}$. We choose unCLIP for its unique hierarchical architecture, which explicitly exposes a \textit{noise-augmented image embedding space}, allowing for precise control over semantic variations. The generation starts from a random Gaussian latent $\rvz_T \sim \mathcal{N}(0, \mathbf{I})$ and proceeds via $T$ denoising steps. We control the synthesis trajectory by modulating two key latent parameters:

\vspace{1mm}\noindent \textbf{\textit{Conditioning Noise Level ($\ell$)}:}
This parameter exploits the noise-conditioning mechanism inherent to unCLIP-style generative training to regulate the strength of semantic perturbations. Specifically, given an input image $\rmI_i$, its conditioning representation is extracted as $\rvc^\text{img}_i = \gV(\rmI_i)$, where $\gV$ denotes the conditioning image encoder (CLIP ViT-H/14). This perturbation occurs in the semantic embedding space. The perturbed embedding $\hat{\rvc}^\text{img}_{i}$ is obtained by injecting Gaussian noise $\boldsymbol{\varepsilon} \sim \mathcal{N}(0, \mathbf{I})$ according to the diffusion schedule at timestep $\ell$:
\begin{equation}
\label{eq:noise_injection}
\hat{\rvc}^\text{img}_{i} = \operatorname{noisy}(\rvc^\text{img}_i, \ell) = \sqrt{\bar{\alpha}_\ell}\rvc^\text{img}_i + \sqrt{1 - \bar{\alpha}_\ell} \boldsymbol{\varepsilon},
\end{equation}
where $\bar{\alpha}_\ell = \prod_{j=1}^{\ell} \alpha_j$ denotes the cumulative product of noise variances in the standard DDPM schedule~\cite{ho2020denoising}. The perturbed embedding $\hat{\rvc}^\text{img}_i$ is then used to guide the reverse diffusion process, generating a new image conditioned on the noised semantics.

\noindent \textit{Effect}: Higher $\ell$ reduces the signal-to-noise ratio of the condition, encouraging the model to "hallucinate" diverse variations (e.g., background, pose) while retaining coarse semantics. Lower $\ell$ ensures high fidelity to the source image.

\vspace{1mm}\noindent \textbf{\textit{Guidance Scale ($g$)}:}
This parameter regulates the strength of text conditioning via Classifier-Free Guidance (CFG)~\cite{ho2022classifier}. Given a caption $\rmT_i$, its semantic representation is obtained as $\rvc^\text{text}_i = \gT(\rmT_i)$, where $\gT$ is the text encoder of the generative model. During each reverse diffusion step $t$, the noise estimate is computed as an interpolation between an unconditional prediction (based on a null caption $\varnothing$) and a text-conditioned one:
\begin{equation}
\epsilon_t = \gG(\rvz_{t}, t, \varnothing) + g \cdot \left( \gG(\rvz_{t}, t, \rvc^\text{text}_i) - \gG(\rvz_{t}, t, \varnothing) \right).
\label{eq:cfg}
\end{equation}
\textit{Effect}: $g$ controls the trade-off between semantic alignment and diversity. Lower $g$ relaxes the textual constraint, fostering creative diversity but risking semantic drift. Higher $g$ enforces strict adherence to the caption, ensuring fidelity but potentially reducing variance.

\subsubsection{Image-Conditioned Adaptive Generation}
\label{method:ada-img}
This mode synthesizes view $\rmI_{i,\text{IC}}$ by performing variations in the conditional image embedding space. We determine the optimal conditioning noise $\ell_i^{\text{ada}}$ through a saliency-aware process.

\vspace{1mm}\noindent\textbf{\textit{Visual Saliency Estimation via Self-Supervised Prior}s.}
We estimate the foreground ratio $p_i \in [0, 1]$ to quantify the semantic stability of an image. Leveraging the \textit{emergent segmentation property} of self-supervised features (as widely observed in ViT-based models like DINO~\cite{caron2021emerging}), we utilize the principal components of feature maps as a robust, training-free saliency proxy.
Specifically, we extract dense features $\rmF_i = \gV(\rmI_i) \in \mathbb{R}^{H \times W \times K}$ using the generator's frozen CLIP image encoder $\gV$.
By projecting $\rmF_i$ onto the first principal component $\rvw$ (computed from a reference set), we obtain the activation map $\rmA_i = \rmF_i \cdot \rvw$. We then apply min-max normalization to obtain $\hat{\rmA}_i$:
\begin{equation}
\hat{\rmA}_i = \frac{\rmA_i - \operatorname{min}(\rmA_i)}{\operatorname{max}(\rmA_i) - \operatorname{min}(\rmA_i)}, \quad \hat{\rmA}_i \in \mathbb{R}^{H \times W}.
\end{equation}
As visualized in Fig.~\ref{fig:framework_GenView++} (a), this unsupervised proxy effectively highlights semantic subjects, distinguishing object-centric images from complex backgrounds without requiring external segmentation labels. The saliency ratio $p_i$ is calculated as the fraction of spatial tokens exceeding a threshold $\alpha$:
\begin{equation}
\label{eq:foreground_proportion}
{p}_i = \frac{1}{HW}\sum_{h=1}^H \sum_{w=1}^W \mathbf{1}\left\{ \hat{\rmA}_{i}[h,w] > \alpha \right\},
\end{equation}
where $\mathbf{1}\{\cdot\}$ is the indicator function. 

\vspace{1mm}\noindent\textbf{\textit{Adaptive Conditioning Noise Injection}.}
Our modulation strategy is grounded in the semantic robustness of visual content. Images with dominant foregrounds (high $p_i$) exhibit a strong semantic core, making them resilient to perturbations in the latent space. For these samples, injecting higher conditioning noise encourages the generator to hallucinate diverse backgrounds while preserving the robust object identity. Conversely, images with low saliency (low $p_i$) often contain scattered or fine-grained details that are fragile to noise; thus, a lower perturbation level is necessary to prevent semantic collapse.

To implement this rigorously, we map the saliency $p_i$ to a discrete noise level $\ell_i^{\text{ada}}$, which corresponds to a specific timestep in the diffusion noise schedule:
\begin{equation}
\begin{aligned}
\ell^\text{ada}_i = \min\left( 400, 100 \cdot \left\lfloor \frac{{p}_i}{0.2} \right\rfloor \right),
\label{eq:l_ada}
\end{aligned}
\end{equation}
where $\ell^\text{ada}_i \in \{0, 100, 200, 300, 400\}$. With the adaptive noise level determined, we apply the general noise injection mechanism defined in Eq.~\ref{eq:noise_injection}. By substituting the specific timestep $\ell = \ell_i^{\text{ada}}$, we obtain the perturbed embedding $\hat{\rvc}_i$ that balances fidelity and diversity for the specific input $\rmI_i$. Finally, the synthetic view is generated by conditioning the diffusion model $\mathcal{G}$ on this embedding:
\begin{equation}
\rmI_{i, \text{IC}} = \mathcal{G}(\rvz_T, \hat{\rvc}_i).
\label{eq:x_ada}
\end{equation}
A random transformation $t_\text{IC}$ is applied to produce the final view $\rmP_{i,\text{IC}} = t_\text{IC}(\rmI_{i,\text{IC}})$.

\subsubsection{Text-Conditioned Adaptive Generation}
\label{method:ada-text}
This mode synthesizes view $\rmI_{i,\text{TC}}$ by generating images solely from textual prompts via the unCLIP text-to-image pathway. We determine the optimal guidance scale $g_i^{\text{ada}}$ through a complexity-aware process.

\vspace{1mm}\noindent\textbf{\textit{Caption Complexity Estimation}.}
We quantify the \textit{semantic information density} of caption $\rmT_i$ into a discrete score $s_i \in \{1,2,3,4\}$. We employ a pretrained language model $\Phi$, prompted with a rule-based scoring instruction $R(\rmT_i)$ (see Table~\ref{supp:rule} for the full prompt):
\begin{equation}
 s_i = \Phi(R(\rmT_i)), \quad s_i \in \{1,2,3,4\}.
 \label{eq:complexity_score}
\end{equation}
A higher score $s_i$ indicates a caption rich in explicit visual constraints (e.g., textures, spatial relations), whereas a lower score implies an abstract or ambiguous description.

\vspace{1mm}\noindent\textbf{\textit{Adaptive Guidance Scale}.}
Our modulation strategy is grounded in the semantic ambiguity of textual descriptions. Simple captions (low $s_i$) inherently suffer from high variance (e.g., "a dog" could map to any breed or background), making them prone to semantic drift. Therefore, they require a higher guidance scale to enforce strict alignment with the prompt. Conversely, complex captions (high $s_i$) already constrain the generative space significantly; in these cases, we relax the guidance (lower $g$) to encourage visual diversity without losing context.
We implement this by mapping the complexity score $s_i$ to the guidance scale $g_i^\text{ada}$ using a linear inverse mapping:
\begin{equation}
g_i^\text{ada}= 10 -2 \cdot s_i, \quad g_i^\text{ada} \in \{2,4,6,8\}.
\label{eq:g_ada}
\end{equation}
Finally, the synthetic view is generated by conditioning the diffusion model $\mathcal{G}$ on the text embedding $\rvc_i^\text{text}$:
\begin{equation}
\rmI_{i,\text{TC}}=\mathcal{G}(\rvz_T, \rvc_i^\text{text}; g_i^\text{ada}).
\end{equation}
A random augmentation $t_\text{TC}$ is applied to produce the final view $\rmP_{i,\text{TC}} = t_\text{TC}(\rmI_{i,\text{IC}})$.

\subsubsection{Image-Text-Conditioned Adaptive Generation}
\label{method:ada-imgtext}
This mode synthesizes view $\rmI_{i,\text{ITC}}$ by leveraging the dual-conditioning capability of Stable unCLIP to bridge the modality gap. We synergistically modulate both latent parameters to achieve fine-grained control.

\vspace{1mm}\noindent\textbf{\textit{Synergistic Parameter Selection}.}
For a given pair $(\rmI_i, \rmT_i)$, we simultaneously apply the adaptive strategies derived from both modalities:
\begin{itemize}
    \item Visual Stream: We compute the adaptive noise level $\ell_i^\text{ada}$ based on the image's foreground saliency (via Eq.~\ref{eq:l_ada}) and obtain the perturbed image embedding $\hat{\rvc}_i$ (via Eq.~\ref{eq:noise_injection}).
    \item Textual Stream: We compute the adaptive guidance scale $g_i^\text{ada}$ based on the caption's complexity (via Eq.~\ref{eq:g_ada}).
\end{itemize}

\vspace{1mm}\noindent\textbf{\textit{Joint-Conditioned Synthesis.}}
The synthetic view is generated by conditioning the diffusion model $\mathcal{G}$ on the joint tuple of the perturbed visual prior and the textual cue:
\begin{equation}
\rmI_{i,\text{ITC}} = \gG(\rvz_i,\ \operatorname{noisy}(\rvc^\text{img}_i, \ell_i^{\text{ada}}),\ \rvc_i^\text{text},\ g_i^{\text{ada}}).
\label{eq:joint_gen}
\end{equation}
By injecting noise into the image embedding ($\hat{\rvc}_i$), we introduce structural variations (e.g., pose, viewpoint). Simultaneously, by anchoring the generation with the text embedding ($\rvc_i^\text{text}$) under adaptive guidance, we ensure these variations remain semantically valid. This dual control effectively mitigates mode collapse (common in image-only variation) and hallucination (common in text-only generation). A random transformation $t_\text{ITC}$ produces the final view $\rmP_{i,\text{ITC}} = t_\text{ITC}(\rmI_{i,\text{ITC}})$.

\subsection{Quality-Driven Contrastive Learning}
\label{method:qua}
Standard contrastive frameworks operate on an agnostic ``augment-and-trust'' assumption, treating every positive pair as equally informative. This is suboptimal for generative augmentation, where synthesized views vary significantly in semantic validity and information content.
To address this, we introduce an online Quality-Driven Contrastive Learning module. As shown in Fig.~\ref{fig:framework_GenView++}~(b), this module functions as a soft denoising curriculum: it explicitly evaluates the quality of each pair using a frozen CLIP assessor and dynamically modulates its contribution to the loss. This ensures the model learns primarily from high-fidelity, high-diversity samples while suppressing noise.

\subsubsection{Intra-Modal Pair Reweighting}
\label{method:qua-II}
For intra-modal pairs, a high-quality view should preserve the foreground object (consistency) while varying the background context (diversity).

\vspace{1mm}\noindent\textbf{\textit{Saliency-Guided Feature Decoupling.}}
For a pair of views $(\rmP_{i,1}, \rmP_{i,2})$, we extract dense visual representations using the frozen CLIP image encoder $\gE^*_{\text{img}}$:
\begin{equation}
 \rmF_{i,1} = \gE^*_\text{img}(\rmP_{i,1}), \quad \rmF_{i,2} = \gE^*_\text{img}(\rmP_{i,2}), \quad \rmF \in \mathbb{R}^{H\times W \times K}.
\end{equation}
To disentangle semantics from context, we again leverage the emergent segmentation property of self-supervised features~\cite{caron2021emerging}. We compute the first principal component of the feature map across channel $K$ and normalize it to obtain the foreground attention map $\rmM_{i}^f \in [0,1]^{H \times W}$. The background map is defined as $\rmM^b_{i} = 1 - \rmM^f_{i}$.
We use these maps to pool the dense features into foreground ($\rvz^f$) and background ($\rvz^b$) vectors:
\begin{equation}
\begin{aligned}
 \rvz^{f}_{i,1} &= \rmM_{i,1}^f \otimes \rmF_{i,1}, \quad \rvz^{f}_{i,2} = \rmM_{i,2}^{f} \otimes \rmF_{i,2}, \\
 \rvz^{b}_{i,1} &= \rmM_{i,1}^b \otimes \rmF_{i,1}, \quad \rvz^{b}_{i,2} = \rmM_{i,2}^{b} \otimes \rmF_{i,2},
\end{aligned}
\end{equation}
where $\otimes$ denotes the spatial pooling operation weighted by the mask: $\rvz = \rmM \otimes \rmF = \sum_{h=1}^{H} \sum_{w=1}^{W} \rmM_{h, w}\rmF_{[h, w, :]}$.

\vspace{1mm}\noindent\textbf{\textit{Quality Scoring and Loss Modulation}.}
We define the pair quality $q_{i}^\text{II}$ by balancing two competing objectives:
\begin{equation}
q_{i}^\text{II} = \underbrace{\operatorname{cos}(\rvz^f_{i,1}, \rvz^f_{i,2})}_{s_i^f \text{ (Fidelity)}} - \underbrace{\operatorname{cos}(\rvz^b_{i,1}, \rvz^b_{i,2})}_{s_i^b \text{ (Redundancy)}}.
\end{equation}

This score penalizes pairs with mismatched objects (low $s_i^f$, likely semantic drift) or identical backgrounds (high $s_i^b$, likely trivial redundancy).
Finally, we normalize these scores via softmax to obtain adaptive weights $w_i^\text{II}$, which modulate the image-image contrastive loss:
\begin{equation}
 w_i^\text{II} = \frac{\exp(q_{i}^\text{II})}{\sum_{j} \exp(q_{j}^\text{II})}, \quad \gL^{\text{II}, \text{QD}}_{i} = w_i^\text{II} \cdot \gL^{\text{II}}_{i}.
\label{eq:loss_quality_II}
\end{equation}

\subsubsection{Cross-Modal Pair Reweighting}
\label{method:qua-IT}
Extending this principle to the multimodal domain, we evaluate a generated view $\rmP_{i,\ast}$ against its caption $\rmT_i$. A high-quality image-text pair must be semantically aligned yet visually novel compared to the original image $\rmI_i$.

\vspace{1mm}\noindent\textbf{\textit{Multimodal Feature Extraction}.}
To ensure valid cross-modal comparisons, we employ a pretrained CLIP model as the metric backbone. We utilize its matched pair of frozen CLIP encoders $(\gE^*_\text{text}, \gE^*_\text{img})$ to project inputs into a shared semantic space. Specifically, we extract the global text embedding $\rve_{i}$ and the dense visual features for both the original and generated images:
\begin{equation}
\begin{aligned}
\rve_{i} &= \gE^*_\text{text}(\rmT_i), \quad \rve_i \in \mathbb{R}^K, \\
\rmF_{i, \text{raw}} &= \gE^*_\text{img}(\rmI_i), \quad \rmF_{i, \ast} = \gE^*_\text{img}(\rmP_{i, \ast}).
\end{aligned}
\end{equation}
Using the background masks from Sec.~\ref{method:qua-II}, we obtain the background context vectors $\rvz_{i, \text{raw}}^b$ and $\rvz_{i,\ast}^b$:
\begin{equation}
\rvz_{i, \text{raw}}^b = \rmM_{i, \text{raw}}^b \otimes \rmF_{i, \text{raw}},\quad
\rvz_{i,\ast}^b = \rmM_{i,\ast}^b \otimes \rmF_{i,\ast}.
\end{equation}

\vspace{1mm}\noindent\textbf{\textit{Quality Score Computation}.}
We compute the cross-modal quality score $q_{i,\ast}^\text{IT}$ based on:
(1) \textbf{Alignment ($s_{i,\ast}^\text{IT}$):} Semantic consistency, measured by the cosine similarity between the text embedding $\rve_i$ and the generated image feature.
(2) \textbf{Redundancy ($s_{i,\ast}^b$):} Visual redundancy, quantified by the background similarity between the generated view and the original image.
\begin{equation}
q_{i,\ast}^\text{IT} = \underbrace{\operatorname{cos}(\rve_i, \operatorname{AvgPool}(\rmF_{i,\ast}))}_{s_{i,\ast}^\text{IT} \text{ (Alignment)}} - \underbrace{\operatorname{cos}(\rvz_{i, \text{raw}}^b, \rvz_{i,*}^b)}_{s_{i,*}^b \text{ (Redundancy)}}.
\end{equation}
This score ensures that we prioritize views that are valid interpretations of the caption (high alignment) but visually distinct from the source (low redundancy).
We normalize these scores across the expanded positive set $\gP_i$ to reweight the image-text loss:
\begin{equation}
w_{i,\ast}^\text{IT} = \frac{\exp(q_{i,\ast}^\text{IT})}{\sum_{j, \ast} \exp(q_{j,\ast}^\text{IT})}, \quad \gL^{\text{IT}, \text{QD}}_{i} = w_{i,\ast}^\text{IT} \cdot \gL^{\text{IT}}_{i}.
\end{equation}
By dynamically down-weighting pairs with low alignment (semantic drift) or high redundancy (trivial positives), GenView++ effectively filters out the noise inherent in generative augmentation, stabilizing the training process.
\section{Main Results}
\label{sec:exp}

\begin{table}[t!]
\centering
\caption{\textbf{Linear classification accuracy on ImageNet-1K.} $^*$: our reproduction.}
\resizebox{\linewidth}{!}{
\begin{tabular}{lccc}
\toprule
Method &Backbone & Epochs & Top-1 \\
\midrule
InstDisc \citet{wu2018unsupervised} &ResNet-50& 200 & 56.5 \\
SimCLR \citet{chen2020simple} &ResNet-50& 200 & 66.8 \\
PCL \citet{li2020prototypical} &ResNet-50& 200 & 67.6 \\
Adco \citet{qi2020learning} &ResNet-50& 200 & 68.6 \\
InfoMin \citet{tian2020makes} &ResNet-50& 200 & 70.1\\
NNCLR \citet{dwibedi2021little} &ResNet-50& 200 & 70.7 \\
LEVEL \citet{huang2022learning} &ResNet-50& 200 & 72.8 \\
Barlow Twins \citet{zbontar2021barlow} &ResNet-50& 300 & 71.4 \\
CLIP \citet{radford2021learning} &ResNet-50& - & 74.3 \\
MoCov2 \citet{he2020momentum} &ResNet-50& 200 & 67.5 \\ 
MoCov2 + C-Crop \citet{peng2022crafting} &ResNet-50& 200 & 67.8\\
\rowcolor{mygray} MoCov2 + GenView++~(V) &ResNet-50& 200 & \textbf{70.0} \\
SwAV \citet{caron2020unsupervised}$^*$ &ResNet-50& 200 & 70.5 \\
\rowcolor{mygray} SwAV + GenView++~(V) &ResNet-50& 200&\textbf{71.7}\\
SimSiam \citet{chen2020exploring} &ResNet-50& 200 & 70.0\\
\rowcolor{mygray} SimSiam + GenView++~(V) &ResNet-50& 200 & \textbf{72.2} \\
BYOL \citet{grill2020bootstrap}$^*$ &ResNet-50& 200 & 71.8 \\
\rowcolor{mygray} BYOL + GenView++~(V) &ResNet-50& 200 & \textbf{73.2} \\ 
MoCov3 \citet{chen2021mocov3} &ResNet-50& 100 & 68.9 \\ % from github
\rowcolor{mygray} MoCov3 + GenView++~(V) &ResNet-50& 100 & \textbf{72.7} \\
MoCov3 \citet{chen2021mocov3} &ResNet-50& 300&72.8\\ % from github
\rowcolor{mygray} MoCov3 + GenView++~(V) &ResNet-50& 300& \textbf{74.8}\\
\midrule
MAE \citet{he2022masked} &ViT-B& 1600 & 68.0 \\
I-JEPA \citet{assran2023self} &ViT-B& 600 & 72.9 \\
VICReg \citet{bardes2022vicregl} &CNX-S& 400 & 76.2 \\
DINO \citet{caron2021emerging}$^*$ &ViT-S& 50 & 70.3 \\
\rowcolor{mygray} DINO + GenView++~(V) &ViT-S& 50 & \textbf{71.9} \\
MoCov3 \citet{chen2021mocov3} &ViT-S& 300 & 73.2\\ % from github
\rowcolor{mygray} MoCov3 + GenView++~(V) &ViT-S& 300 & \textbf{74.5}\\
MoCov3 \citet{chen2021mocov3} &ViT-B& 300 & 76.7 \\ % from github
\rowcolor{mygray} MoCov3 + GenView++~(V) &ViT-B& 300 & \textbf{77.8} \\
\bottomrule
\end{tabular}
}
\label{tab:exp_linear}
\end{table}

\subsection{Vision Representation Benchmarks}
\label{exp:vision}
We integrate GenView++~(V) into six representative SSL frameworks: MoCov2~\cite{he2020momentum}, BYOL~\cite{grill2020bootstrap}, DINO~\cite{caron2021emerging}, SwAV~\cite{caron2020unsupervised}, SimSiam~\cite{chen2020exploring}, and MoCov3~\cite{chen2021mocov3} using ResNet and ViT backbones.

\subsubsection{Linear Classification on ImageNet}
Table~\ref{tab:exp_linear} presents the linear probing results on ImageNet-1K. GenView++ consistently enhances representation quality across diverse SSL paradigms, including contrastive (e.g., MoCov2, MoCov3) and non-contrastive (e.g., DINO, SimSiam, BYOL, SwAV) approaches. It demonstrates broad applicability and effectiveness with both ResNet and ViT backbones.
Notable observations include:

\vspace{1mm}\noindent \textbf{(1) Universality:} Improvements are consistent across both CNN (ResNet-50) and Transformer (ViT) backbones, demonstrating that our generative augmentation strategy captures fundamental semantic invariants independent of the architectural inductive bias.

\vspace{1mm}\noindent \textbf{(2) Data Efficiency:} When integrated with MoCov3 (300 epochs), GenView++ achieves \textbf{74.8\%} top-1 accuracy, outperforming CLIP (74.3\%) despite using orders of magnitude fewer data pairs (1.28M vs. 400M). This validates the high information density of our adaptively generated views.

\vspace{1mm}\noindent \textbf{(3) Superiority over Handcrafted Baselines:} Compared to C-Crop~\cite{peng2022crafting}, which optimizes geometric cropping, GenView++ yields larger gains (+2.2\% on MoCoV2). This confirms that diffusion-based synthesis introduces semantic-level diversity (e.g., novel poses, backgrounds) that geometric transformations cannot achieve.

\subsubsection{Transfer Learning to Dense Prediction}

\begin{table}[t!]
\centering
\caption{\textbf{Transfer learning performance on MS-COCO for object detection and instance segmentation.} $^*$: our reproduction.}
\label{tab:exp_coco}
\resizebox{\linewidth}{!}{
\begin{tabular}{@{}lcccccc@{}}
\toprule
\multirow{2}{*}{Method} & \multicolumn{3}{c}{Object Det.} & \multicolumn{3}{c}{Instance Seg.} \\
\cmidrule(l{3pt}r{3pt}){2-4} \cmidrule(l{3pt}r{3pt}){5-7}
& AP & AP$_{50}$ & AP$_{75}$ & AP & AP$_{50}$ & AP$_{75}$ \\
\midrule
ReSim \citet{xiao2021region} & 39.8 & 60.2 & 43.5 & 36.0 & 57.1 & 38.6 \\
DenseCL \citet{wang2021dense} & 40.3 & 59.9 & 44.3 & 36.4 & 57.0 & 39.2 \\ \midrule
SimSiam \citet{chen2020exploring}$^*$& 38.5 & 57.8 & 42.3 & 34.7 & 54.9 & 37.1 \\
\rowcolor{mygray} SimSiam + GenView++~(V)& \textbf{39.1} & \textbf{58.5} & \textbf{43.0} & \textbf{35.2}& \textbf{55.9} & \textbf{37.7}\\ \midrule
MoCov2 \citet{chen2020improved}$^*$ & 39.7 & 59.4 & 43.6 & 35.8 & 56.5 & 38.4 \\ 
MoCov2 + FreeATM \citet{zhang2023free} & 40.1& -& -& -& -&- \\ 
\rowcolor{mygray} MoCov2 + GenView++~(V) & \textbf{40.5} & \textbf{60.0} & \textbf{44.3} & \textbf{36.3} & \textbf{57.1} & \textbf{38.9} \\\midrule
BYOL \citet{grill2020bootstrap}$^*$ & 40.6 & 60.9 & 44.5 & 36.7 & 58.0 & 39.4 \\
\rowcolor{mygray} BYOL + GenView++~(V) & \textbf{41.2} & \textbf{61.5} & \textbf{44.9} & \textbf{37.0} & \textbf{58.4} & \textbf{39.7} \\
\bottomrule
\end{tabular}
}
\vspace{-4mm}
\end{table}

To assess the spatial granularity of learned features, we evaluate transfer performance on MS-COCO~\cite{lin2014microsoft} object detection and instance segmentation (Table~\ref{tab:exp_coco}).
GenView++ consistently improves box AP and mask AP across all baselines. Crucially, it outperforms FreeATM~\cite{zhang2023free}, a text-prompted augmentation method. This indicates that our image-conditioned adaptive strategy (saliency-guided noise injection) effectively preserves fine-grained spatial structures essential for pixel-level tasks, avoiding the semantic drift often seen in purely text-driven generation.

\subsubsection{Data Efficiency Analysis}
\label{supp:naive}

We explicitly investigate whether the performance gains of GenView++ stem merely from increased data volume or from the superior quality of our adaptive views. To disentangle these factors, we conduct a rigorous comparative study using MoCo v3 (ResNet-50, 50 epochs) against three distinct expansion strategies:
\begin{itemize}
    \item \textbf{Web Retrieval (+Laion400M):} To simulate expanding with noisy web data while minimizing domain shift, we use a retrieval-based technique~\cite{beaumont-2022-clip-retrieval} to select 0.3M images from Laion400M that are semantically nearest to IN1K samples.
    \item \textbf{Real Expansion (+IN21K):} We randomly sample 0.3M real images from ImageNet-21K with labels matching the IN1K classes. This represents the "gold standard" of adding real, in-domain data.
    \item \textbf{Unguided Synthesis (+Synthetic Naive):} We generate 0.3M synthetic images without our adaptive noise/guidance control, representing a baseline generative augmentation.
    \item \textbf{Ours (+GenView++):} We use only 0.15M adaptively generated views (half the volume of baselines).
\end{itemize}

\begin{table}[t!]
\caption{\textbf{Data efficiency analysis on ImageNet-1K.} Comparison against external data expansion and naive generation strategies. Note that GenView++ uses only \textbf{half} the number of additional samples compared to baselines.}
\label{tab:exp_naive}
\begin{tabular}{lccc}
\toprule
Dataset & Added Images & Top-1 & Top-5 \\ \midrule
Baseline (IN1K) & - & 62.39 & 84.57 \\
+ Laion400M & 0.3M (Retrieved) & 63.31 & 85.53 \\
+ IN21K & 0.3M (Real) & 64.10 & 85.86 \\
+ Synthetic (Naive) & 0.3M (Generated) & 63.36 & 85.14 \\
\rowcolor{mygray} + GenView++~(V) & \textbf{0.15M (Adaptive)} & \textbf{65.62} & \textbf{87.25} \\
\bottomrule
\end{tabular}
\end{table}

As shown in Table~\ref{tab:exp_naive}, expanding with Laion400M or Naive Synthetic images yields only marginal gains (+0.92\% and +0.97\%). This suggests that simply increasing data volume is inefficient if the additional samples suffer from \textit{domain gaps} (Laion) or \textit{semantic drift} (Naive Synthetic).
Expanding with real IN21K data brings moderate improvements (+1.71\%) due to the high distributional alignment. Crucially, GenView++ achieves the highest accuracy (65.62\%), outperforming the IN21K expansion by a significant margin (\textbf{+1.52\%}), despite using \textbf{50\% fewer added samples} (0.15M vs. 0.3M).
This result validates our core premise: \textit{data quality matters more than quantity}. Our adaptive generation strategy constructs "hard positive" pairs that lie closer to the decision boundary than random real-world samples. These views possess higher information density, providing stronger gradients for representation learning and justifying the offline generation cost through superior sample efficiency.

\subsubsection{Comparison with View Construction Methods}
To verify the Efficacy of GenView++~(V) against state-of-the-art positive pair construction methods, we conduct comprehensive pretraining and linear evaluation on CIFAR-10 \citet{krizhevsky2009learning} (CF10), CIFAR-100 \citet{krizhevsky2009learning} (CF100), and Tiny ImageNet \citet{le2015tiny} (TinyIN).
We analyze performance across three distinct paradigms of view construction:
\begin{itemize}
    \item \textbf{Geometric Transformation:} Methods like \textbf{C-Crop} \citet{peng2022crafting} optimize spatial transformations. While effective, they are bounded by the pixel information present in the single instance.
    \item \textbf{Manifold Interpolation:} Methods such as \textbf{ViewMaker}~\cite{tamkin2020viewmaker}, \textbf{NTN}~\cite{kim2023neural}, and \textbf{Local Manifold Augmentation (LMA)}~\cite{yang2022local} synthesize views by mixing or perturbing features based on the existing dataset distribution. They essentially interpolate within the convex hull of observed data.
    \item \textbf{Generative Extrapolation:} Methods like \textbf{DiffAug}~\cite{zang2023boosting} and \textbf{$\mathcal{W}$-perturb}~\cite{han2023constructive} introduce external semantic diversity. GenView++ belongs to this category but distinguishes itself via \textit{adaptive controllability}.
\end{itemize}

\begin{table}[t!]
\centering
\caption{\textbf{Linear classification results of view construction methods on CIFAR-10, CIFAR-100, and TinyImageNet.} We categorize baselines by the nature of variance they introduce.}
\label{tab:exp_tiny}
\resizebox{\linewidth}{!}{
\begin{tabular}{lccc}
\toprule
Method & CF10& CF100& TinyIN\\ \midrule
\multicolumn{4}{c}{\emph{Geometric Transformation (Within-Instance)}}\\
MoCov2 + C-Crop \citet{peng2022crafting}& 88.78& 57.65& 47.98\\
BYOL + C-Crop \citet{peng2022crafting}& 92.54& 64.62 &47.23\\
\midrule
\multicolumn{4}{c}{\emph{Manifold Interpolation (Within-Dataset)}}\\
SimCLR + ViewMaker \citet{tamkin2020viewmaker}& 86.30& -& -\\
SimCLR + NTN \citet{kim2023neural} & 86.90 & - & - \\
MoCov2 + LMA \citet{yang2022local} & 92.02 & 64.89 & - \\
SimSiam + LMA \citet{yang2022local} & 92.46 & 65.70 & - \\
Simsiam + DiffAug \citet{zang2023boosting} & 87.30& 60.10& 45.30 \\
\midrule
\multicolumn{4}{c}{\emph{Generative Extrapolation (Beyond-Dataset)}}\\
W-perturb \citet{han2023constructive}& 92.90 & -& 51.05 \\
\rowcolor{mygray} MoCov2 + GenView++~(V) & 93.00 & 67.49 & \textbf{56.76}\\ 
\rowcolor{mygray} BYOL + GenView++~(V) & \textbf{93.56} & \textbf{67.53} & 54.79 \\ 
\bottomrule
\end{tabular}
}
\vspace{-2mm}
\end{table}

\begin{table*}[th!]
\centering
\caption{\textbf{Linear classification accuracy on visual benchmarks.} ``I-I" and ``I-T" indicate the number of image-image and image-text pairs used. StableRep results are reproduced by us with official weigth!s.}
\resizebox{\linewidth}{!}{
\begin{tabular}{ll|cc|cccccccccc|c}
\toprule
Method & Venue & I-I& I-T & IN1K & CF10 & CF100 & Aircraft & DTD & Flowers & Pets & SUN397 & Caltech-101 & Food-101 &Avg.\\
\midrule
CLIP \citet{radford2021learning} & ICML'21 & 0&1 & 53.30& 81.80& 62.70& 34.70& 57.30& 84.10& 60.50& 54.30& 75.60& 58.70&62.30\\
\rowcolor{mygray} GenView++~(VL) & - & 0&1 & \textbf{60.49}& \textbf{92.24}& \textbf{76.08}& \textbf{35.01}& \textbf{64.47}& \textbf{86.18}& \textbf{69.96}& \textbf{62.41}& \textbf{92.45}& \textbf{68.24}&\textbf{70.75}\\
\midrule
StableRep \citet{tian2023stablerep} & NeurIPS'23 & 6&0 & 63.86& 93.03& 76.02& 29.19& \textbf{69.84}& 83.53& 70.51& \textbf{65.53}& 90.67& 70.91&71.31\\
\rowcolor{mygray} GenView++~(VL) & - & 6&0 & \textbf{64.38}& \textbf{93.53}& \textbf{77.33}& \textbf{32.64}& 69.15& \textbf{86.86}& \textbf{73.04}& 65.77& \textbf{91.47}& \textbf{72.39}&\textbf{72.66}\\
\midrule
SLIP \citet{mu2022slip} & ECCV'22 & 1&1 & 65.40& - & - & - & - & - & - & - & - & - & -\\
\rowcolor{mygray} GenView++~(VL) & - & 1&1 & \textbf{66.44}& 94.53& 79.23& 38.46& 69.47& 87.06& 75.88& 67.43& 91.99& 74.13&74.46\\
\rowcolor{mygray} GenView++~(VL)& - & 6&4& \textbf{68.17}& \textbf{95.08}& \textbf{80.26}& \textbf{40.23}& \textbf{71.17}& \textbf{90.00}& \textbf{77.46}& \textbf{69.03}& \textbf{92.22}& \textbf{75.96}&\textbf{75.96}\\
\bottomrule
\end{tabular}
}
\label{tab:exp_vl_linear}
\end{table*}

\begin{table*}[th!]
\centering
\caption{\textbf{Zero-shot classification results on various downstream datasets.}
Results for CLIP, SLIP, and TripletCLIP are reproduced by us using their official pretrained weigth!s.}
\resizebox{\linewidth}{!}{
\begin{tabular}{ll|cccccccccc|c}
\toprule
Method& Venue& IN1K& CF10& CF100& Aircraft& DTD& Flowers& Pets& SUN397& Caltech-101& Food-101& Avg.\\
\midrule
CLIP \citet{radford2021learning}& ICML'21& 16.34& 41.10& 17.42& 1.11& 11.49& 10.25& 11.37& 32.17& 49.66& 10.73& 20.16\\
SLIP \citet{mu2022slip}& ECCV'22& 23.00& 64.38& 33.61& 1.23& 13.19& 13.58& \textbf{16.68}& 33.00& 58.16& \textbf{14.77}& 27.16\\
LaCLIP \citet{fan2023improving}& NeurIPS'23& 21.50& 57.10& 27.50& \textbf{1.60}& 16.60& \textbf{14.70}& 15.60& 35.10& 52.70& 14.20& 25.66\\
TripletCLIP \citet{patel2024tripletclip}& NeurIPS'24& 7.00& 25.26& 12.51& 1.29& 10.69& 6.34& 5.51& 14.78& 31.54& 4.84& 5.09\\
\rowcolor{mygray} GenView++~(VL) & -& \textbf{23.16}& \textbf{87.82}& \textbf{50.64}& 1.59& \textbf{19.10}& 12.50& 12.73& \textbf{42.44}& \textbf{60.63}& 14.07& \textbf{32.47}\\
\bottomrule
\end{tabular}
}
\label{tab:exp_zeroshot-cls}
\end{table*}
As presented in Table~\ref{tab:exp_tiny}, GenView++ demonstrates superior generalization capability:

\noindent \textbf{(1) Extrapolation vs. Interpolation:} GenView++ consistently outperforms interpolation-based methods (e.g., LMA, Mixup). On CIFAR-100, MoCov2 + GenView++ surpasses MoCov2 + LMA by \textbf{+2.60\%} (67.49\% vs. 64.89\%). This confirms that injecting rich, external visual priors via diffusion models effectively \textit{expands the training manifold} beyond the limits of the original dataset, providing more discriminative features than merely manipulating internal statistics.

\noindent \textbf{(2) Controlled vs. Uncontrolled Generation:} GenView++ outperforms other generative approaches like $\mathcal{W}$-perturb (e.g., \textbf{+5.71\%} on TinyIN). Standard generative methods often inject noise indiscriminately, risking semantic drift. In contrast, our saliency-guided adaptive strategy ensures that high-noise perturbations are restricted to robust foregrounds, while fragile scenes receive lower noise. This mechanism maintains high \textbf{semantic fidelity}, preventing the generated views from acting as noisy labels that corrupt the representation learning process.

\begin{table*}[t!]
\centering
\caption{\textbf{
Zero-shot cross-modal retrieval performance on MS COCO, Flickr30k, and Flickr8k.} We report Recall@1 (R@1) and Recall@5 (R@5) for Text-to-Image and Image-to-Text retrieval. CLIP and SLIP results are reproduced by us.}
\resizebox{\linewidth}{!}{
\begin{tabular}{ll|cccc|cccc|cc}
\toprule
\multirow{3}{*}{Method} & \multirow{3}{*}{Venue} & 
\multicolumn{4}{c}{MS Coco} & \multicolumn{4}{c}{Flickr30k} & \multicolumn{2}{c}{Flickr8k} \\
& & \multicolumn{2}{c}{Text$\rightarrow$Image} & \multicolumn{2}{c}{Image$\rightarrow$Text}
& \multicolumn{2}{c}{Text$\rightarrow$Image} & \multicolumn{2}{c}{Image$\rightarrow$Text}
& \multicolumn{1}{c}{Text$\rightarrow$Image} & \multicolumn{1}{c}{Image$\rightarrow$Text} \\
\cmidrule(lr){3-4} \cmidrule(lr){5-6} \cmidrule(lr){7-8} \cmidrule(lr){9-10} \cmidrule(lr){11-11} \cmidrule(lr){12-12}
& & R@1 & R@5 & R@1 & R@5 & R@1 & R@5 & R@1 & R@5 & R@1 & R@1 \\
\midrule
CLIP \citet{radford2021learning} & ICML'21 & 9.80 & 25.78 & 13.40 & 31.00 & 19.12 & 41.32 & 25.80 & 54.50 & 20.00 & 29.00 \\
SLIP \citet{mu2022slip} & ECCV'22 & 15.25 & 34.87 & \textbf{20.60} & 43.88 & 27.74 & 51.48 & 36.90 & 63.80 & 28.36 & 40.10 \\
TripletCLIP \citet{patel2024tripletclip} & NeurIPS'24 & - & 11.28 & - & 10.38 & - & 22.00 & - & 22.00 & - & - \\
\rowcolor{mygray} GenView++~(VL) & - & \textbf{16.12} & \textbf{36.80} & 20.12 & \textbf{45.02} & \textbf{28.16} & \textbf{54.82} & \textbf{38.20} & \textbf{67.00} & \textbf{29.08} & \textbf{41.10} \\
\bottomrule
\end{tabular}
}
\label{tab:exp_retreival}
\end{table*}

\subsection{Vision-Language Representation Benchmarks}
\label{exp:vision_language}
We evaluate GenView++~(VL) on linear probing, zero-shot classification, and cross-modal retrieval. We use a ViT-B/16 backbone pretrained on CC3M following StableRep \citet{tian2023stablerep} protocols.

\subsubsection{Linear Classification}
Table~\ref{tab:exp_vl_linear} quantifies the transferability of learned representations across 10 diverse downstream benchmarks. We analyze the performance under three distinct settings to isolate the contributions of our framework's components.

\vspace{1mm}\noindent \textbf{(1) Quality vs. Quantity (0 I-I, 1 I-T):}
To strictly evaluate the efficacy of our Quality-Driven Contrastive Learning module, we restrict the input to a single image-text pair per sample, identical to the computational budget of CLIP.
GenView++ achieves an average accuracy of \textbf{70.75\%}, surpassing CLIP (62.30\%) by a substantial \textbf{+8.45\%}.
Since the view budget is identical, this gain cannot be attributed to data augmentation. Instead, it conclusively proves that our dynamic reweigth!ing mechanism effectively allows the model to learn robust features from the same raw data by filtering out misaligned or noisy pairs that degrade standard contrastive training.

\vspace{1mm}\noindent \textbf{(2) Multi-Source vs. Text-Only (6 I-I, 0 I-T):}
We compare our Multi-Source Adaptive Generation against StableRep~\cite{tian2023stablerep}, which relies solely on text-conditioned synthesis. Here, we utilize image-image pairs from our generated pool without text supervision.
GenView++ outperforms StableRep on 9 out of 10 datasets. The advantage is most pronounced in fine-grained recognition tasks, such as Aircraft (\textbf{+3.45\%}).
Text-only generation often suffers from \textit{structural loss} (e.g., losing the specific design details of an aircraft variant). Our IC mode addresses this by injecting noise into the visual embedding, thereby preserving fine-grained structural constraints while diversifying the background. This confirms the necessity of multi-source conditioning for fine-grained representation learning.

\vspace{1mm}\noindent \textbf{(3) Unified Synergy (Full GenView++):}
Finally, we evaluate the full potential of the framework by combining all adaptive views (6 I-I, 4 I-T).
This setting yields state-of-the-art performance, reaching \textbf{68.17\%} on ImageNet-1K and \textbf{75.96\%} on Food-101.
While this configuration incurs a higher computational cost per iteration, it demonstrates the complementary nature of our approach: intra-modal views provide visual invariance (robustness to augmentation), while cross-modal views provide semantic grounding (alignment with language). The synergistic optimization of both objectives drives the representation quality to new heigth!s.

\subsubsection{Zero-Shot Classification}
As shown in Table~\ref{tab:exp_zeroshot-cls}, GenView++ demonstrates superior generalization, achieving an average accuracy of \textbf{32.47\%} and outperforming CLIP (\textbf{+12.31\%}) and LaCLIP (\textbf{+6.81\%}). Crucially, unlike methods such as TripletCLIP~\cite{patel2024tripletclip} that rely on explicit hard negative mining, GenView++ enhances performance by \textbf{enriching the positive signal}. By synthesizing diverse yet semantically consistent positives, we implicitly sharpen the decision boundaries against negatives. This strategy yields robust transferability, particularly on general-purpose benchmarks like CIFAR-10.

\subsubsection{Zero-Shot Cross-Modal Retrieval}
Table~\ref{tab:exp_retreival} evaluates the alignment of the learned visual and textual manifolds. GenView++ consistently surpasses strong baselines, exceeding SLIP by \textbf{+3.20\%} on Flickr30k (Image-to-Text R@5) and \textbf{+1.93\%} on MS COCO (Text-to-Image R@5). These gains confirm that our quality-driven contrastive mechanism effectively filters out mismatched pairs, creating a highly aligned joint embedding space that facilitates precise cross-modal matching without task-specific fine-tuning.

\begin{table}[t!]
\centering
\caption{\textbf{Effect of core components in GenView++~(V) under linear evaluation on IN100.} ``Our framework'' denotes real-to-synthetic view construction without adaptive noise control or quality-driven loss. ``AdaGen'' refers to adaptive image-conditioned view generation. ``Qual.Driv.Cont'' indicates the use of quality-driven contrastive loss.}
\label{tab:exp:component}
\resizebox{\linewidth}{!}{
\begin{tabular}{cccc}
\toprule
Our framework & AdaGen & Qual.Driv.Cont & Top-1 \\
\midrule
$\times$ & $\times$ & $\times$ & 65.52\\
$\times$ & $\times$ & $\checkmark$ & 66.97 ($\uparrow$ 1.45)\\
$\checkmark$ & $\times$ & $\times$ & 71.50 ($\uparrow$ 5.98)\\
$\checkmark$ & $\checkmark$ & $\times$ & 73.96 ($\uparrow$ 8.44)\\
$\checkmark$ & $\times$ & $\checkmark$ & 74.88 ($\uparrow$ 9.36)\\
\rowcolor{mygray}$\checkmark$ & $\checkmark$ & $\checkmark$ & \textbf{75.40 ($\uparrow$ 9.88)}\\
\bottomrule
\end{tabular}
}
\end{table}

\section{Ablation Studies and Analysis}

\subsection{Vision Component Ablation}
\label{exp:ablation_vision}
We analyze the contribution of each unimodal component using ResNet-18 on ImageNet-100.
Table~\ref{tab:exp:component} shows that the base generative framework (Real-to-Synthetic pairing) alone yields a significant \textbf{+5.98\%} gain.
Refining this with \textbf{Adaptive Generation (AdaGen)} boosts accuracy to 73.96\% (\textbf{+8.44\%} over baseline), validating the importance of saliency-aware noise modulation.
Finally, integrating \textbf{Quality-Driven Contrastive Learning} achieves the best performance (\textbf{75.40\%}, \textbf{+9.88\%} total gain), demonstrating the strong synergy between controllable view synthesis and quality-aware supervision.

\subsection{Vision-Language Component Ablation}
\label{exp:ablation_vision_language}
We analyze multimodal components using ViT-B/16 pretrained on a 1M subset of CC3M, with enhancements from our adaptive view generation module.

\begin{table}[t!]
\centering
\caption{\textbf{Efficacy of multi-source adaptive generation.} We compare different generation sources (IC, TC, ITC) and the effect of our adaptive strategy (AdaGen).}
\label{tab:exp_ablation_adagen}
\resizebox{\linewidth}{!}{
\begin{tabular}{cccccccc}
\toprule
\multicolumn{4}{c}{Generation Setting} & \multicolumn{2}{c}{500k Aug.} & \multicolumn{2}{c}{100k Aug.} \\
\cmidrule(lr){1-4} \cmidrule(lr){5-6} \cmidrule(lr){7-8}
IC & TC & ITC & AdaGen & CF10 & IN100 & CF10 & IN100 \\
\midrule 
$\times$ & $\times$ & $\times$ & $\times$ & 83.41 & 66.76 & 83.41 & 66.76 \\ \midrule
$\checkmark$ & $\times$ & $\times$ & $\times$ & 86.98 & 72.96 & 85.17 & 68.98 \\
$\checkmark$ & $\times$ & $\times$ & $\checkmark$ & 88.05 & 76.50 & 86.02 & 71.54 \\ \midrule
$\times$ & $\checkmark$ & $\times$ & $\times$ & 87.55 & 72.60 & 84.84 & 69.14 \\
$\times$ & $\checkmark$ & $\times$ & $\checkmark$ & 88.99 & 75.26 & 85.62 & 70.86 \\ \midrule
$\times$ & $\times$ & $\checkmark$ & $\times$ & 87.39 & 72.86 & 84.40 & 68.92 \\
$\times$ & $\times$ & $\checkmark$ & $\checkmark$ & 88.60 & 76.38 & 85.45 & 71.40 \\ \midrule
\rowcolor{mygray}$\checkmark$ & $\checkmark$ & $\checkmark$ & $\checkmark$ & 90.69 & 78.38 & 87.61 & 74.28 \\
\bottomrule
\end{tabular}
}
\end{table}

\subsubsection{Efficacy of Multi-Source Adaptive Generation}
Table~\ref{tab:exp_ablation_adagen} validates the hierarchical benefits of our generation framework:

\vspace{1mm}\noindent \textbf{(1) Baseline Enrichment (Single Source):}
Introducing any generative source (IC, TC, or ITC) consistently outperforms the baseline. For instance, adding IC views (500k budget) boosts ImageNet-100 accuracy by \textbf{+6.20\%}. This confirms that simply expanding the training distribution with synthetic samples—even without adaptive control—provides valuable regularization.

\vspace{1mm}\noindent \textbf{(2) Controllability Gain (AdaGen).}
Activating our Adaptive Generation mechanism yields further improvements across all modes. For IC, AdaGen lifts performance from 72.96\% to \textbf{76.50\%}. Similar gains are observed for TC and ITC modes. This empirically proves that our saliency-guided noise injection and complexity-aware guidance effectively filter out semantic drift, ensuring that the generated diversity is actually constructive for learning.

\vspace{1mm}\noindent \textbf{(3) Synergistic Complementarity (Unified).}
The full synergy of all three modes achieves the highest performance (\textbf{78.38\%} on IN100). This demonstrates that the different generation strategies are not redundant but complementary: IC mode provides fine-grained visual structural variations, while TC mode offers high-level semantic abstractions. Their combination yields a maximally informative representation space.

\begin{table}[t!]
\centering
\caption{
\textbf{Efficacy of quality-driven contrastive learning.}
``Qual.Driv.Cont" denotes quality-driven loss, and ``Lang.Sup" denotes the image-text contrastive loss.}
\label{tab:exp_ablation_qd}
\resizebox{\linewidth}{!}{
\begin{tabular}{cccccc}
\toprule
\multirow{2}{*}{Qual.Driv.Cont} & \multirow{2}{*}{Lang.Sup} & \multicolumn{2}{c}{500k Aug.} & \multicolumn{2}{c}{100k Aug.} \\
\cmidrule(lr){3-4} \cmidrule(lr){5-6}
& & CF10 & IN100 & CF10 & IN100 \\
\midrule 
$\times$ & $\times$ & 90.69 & 78.38 & 87.61 & 74.28 \\
$\checkmark$ & $\times$ & 91.61 & 79.58 & 90.77 & 78.10 \\
$\times$ & $\checkmark$ & 92.03 & 80.04 & 91.98 & 79.64 \\
\rowcolor{mygray}$\checkmark$ & $\checkmark$ & 92.85 & 81.38 & 92.72 & 80.56 \\
\bottomrule
\end{tabular}
}
\end{table}

\subsubsection{Efficacy of Quality-Driven Contrastive Learning}
Table~\ref{tab:exp_ablation_qd} isolates the contribution of our supervision mechanism. 
\textbf{Vision-Only Setting:} Without textual guidance, generative views are prone to semantic drift. Here, our quality-driven loss acts as a critical \textbf{denoising filter}, boosting accuracy by \textbf{+3.82\%}. This proves that dynamically down-weigth!ing low-fidelity pairs is essential for stabilizing training on synthetic data.
\textbf{Vision-Language Setting:} The benefit persists in the multimodal context (\textbf{+1.34\%}), confirming that our cross-modal alignment metric successfully prioritizes well-aligned pairs, refining the joint embedding space beyond standard symmetric losses.

\subsubsection{Scalability with Generative Budget}
We analyze the scalability of GenView++ by varying the synthetic data volume (100k vs. 500k). As shown in Tables~\ref{tab:exp_ablation_adagen} and~\ref{tab:exp_ablation_qd}, performance improves monotonically with data scale (e.g., a further \textbf{+0.82\%} gain in the full setting). This indicates that our framework effectively capitalizes on the expanded \textbf{semantic diversity} of the larger generative pool without saturating or overfitting to noise, validating the robustness of our adaptive control mechanism.

\subsection{Visualization of Adaptive Generation}

\begin{figure}[t!]
\centering
\includegraphics[width=1\linewidth]{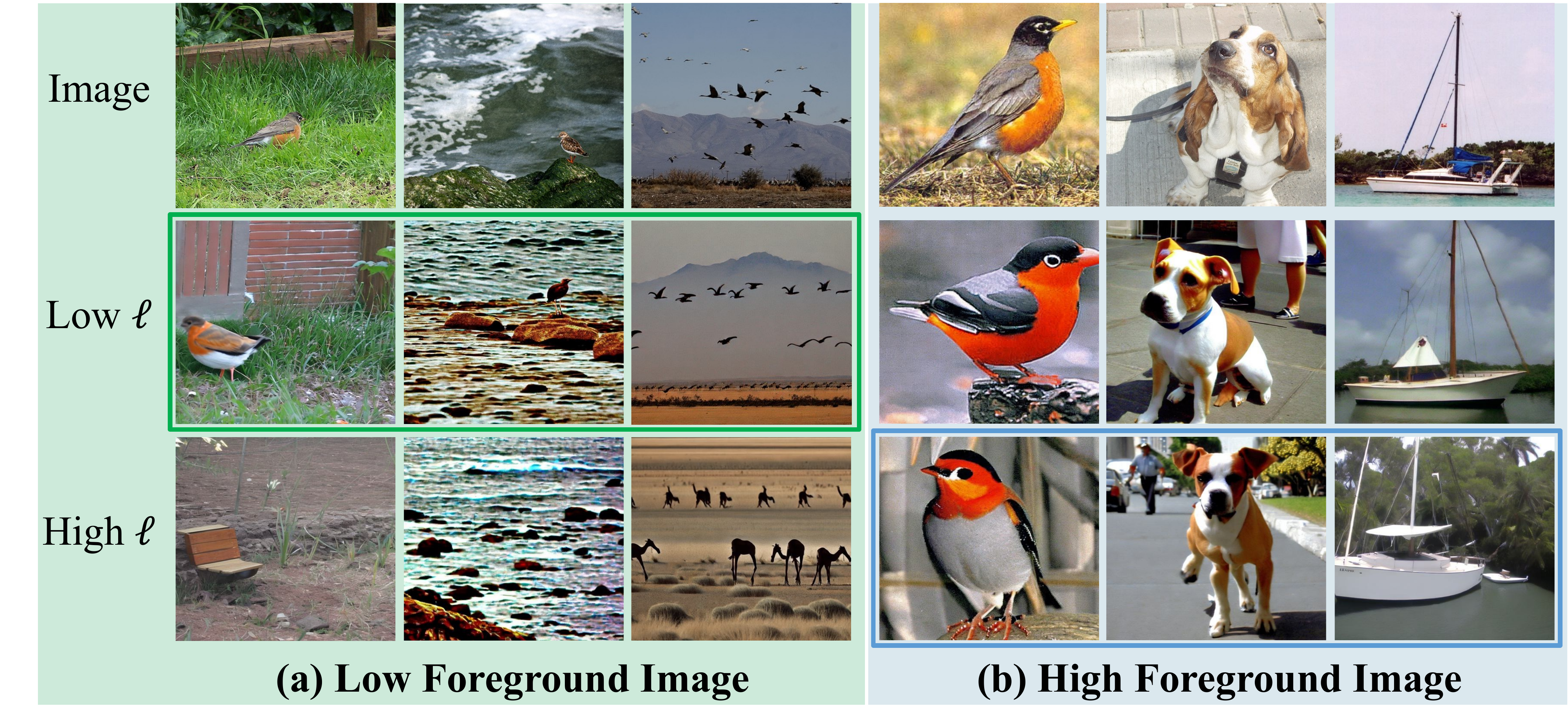}
\caption{\textbf{Image-Conditioned Adaptive Generation.} The noise level $\ell$ is adapted to the image’s foreground proportion. (a) For low-foreground images, a low $\ell$ (green boxes) avoids semantic drift, object loss, or distortion (Col 1–3). (b) For high-foreground images, a high $\ell$ (blue boxes) enriches diversity, e.g., varying pose (Col 4), action (Col 5), and background (Col 6).}
\label{fig:adaptive_noise_level}
\vspace{-4mm}
\end{figure}

\begin{figure*}[th!]
\centering
\includegraphics[width=1\linewidth]{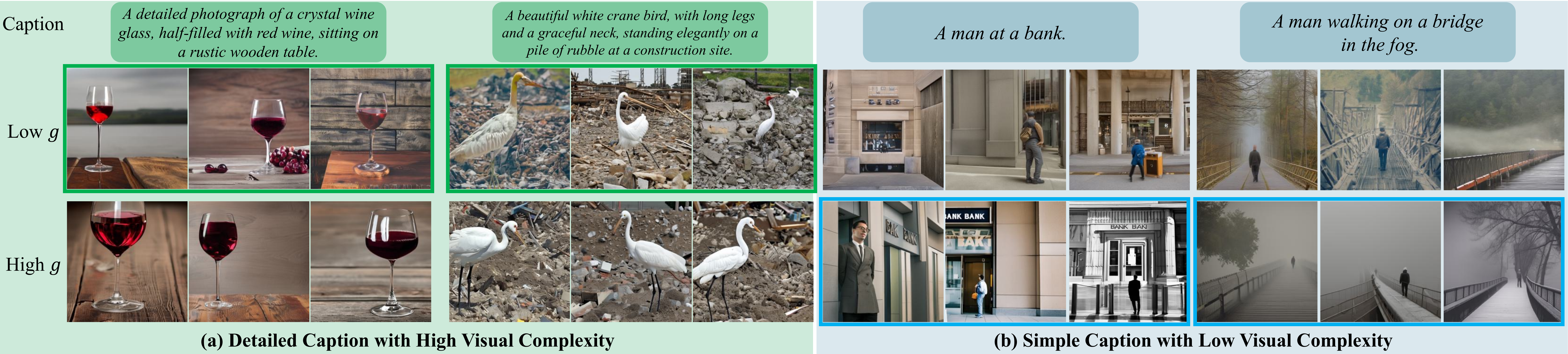}
\caption{\textbf{Text-Conditioned Adaptive Generation.} The guidance scale $g$ is adjusted by caption complexity. (a) Detailed captions with high visual complexity: a low $g$ (green boxes) encourages diverse generations while preserving fine semantics. In contrast, a high $g$ over-constrains the output, yielding repetitive or rigid results. (b) Simple captions with low visual complexity: a high $g$ (blue boxes) strengthens text-image alignment and ensures key concepts are retained.}
\label{fig:adaptive_guidance_scale}
\end{figure*}

\begin{figure*}[th!]
\centering
\includegraphics[width=1\linewidth]{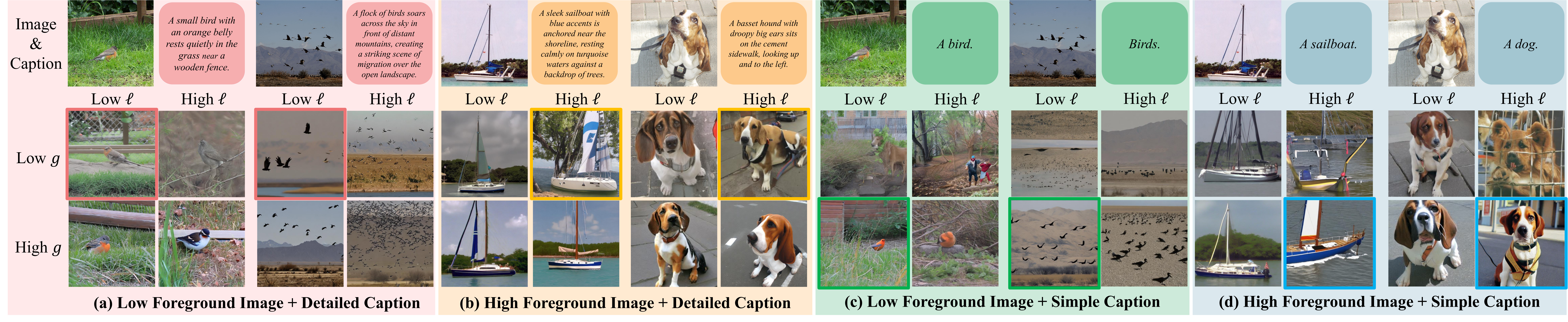}
\caption{\textbf{Image-Text-Conditioned Adaptive Generation.}
This mode jointly adapts noise level $\ell$ and guidance scale $g$ for fine-grained cross-modal control. 
(a) Low-foreground + detailed captions: low $\ell$ preserves small subjects and low $g$ promotes diverse poses and environments (pink boxes); high $\ell$ or $g$ leads to semantically mismatches (e.g., Row 2, Col 2) or reduced diversity (e.g., Row 3, Col 1).
(b) High-foreground + detailed captions: high $\ell$ enriches variation and low $g$ avoids over-constraining by text (yellow boxes); low $\ell$ or high $g$ limits diversity. 
(c) Low-foreground + simple captions: low $\ell$ preserves subjects while high $g$ enforces text grounding (green boxes); low $g$ induces weak textual constraints, risking semantic drift from the intended concept (e.g., unrelated humans in Row 2, Col 10).
(d) High-foreground + simple captions: jointly high $\ell$ and $g$ produce diverse yet consistent generations grounded in strong visual anchors and minimal text.}
\label{fig:adaptive_ITC}
\end{figure*}

Figures~\ref{fig:adaptive_noise_level}, \ref{fig:adaptive_guidance_scale}, and \ref{fig:adaptive_ITC} provide qualitative validation of our adaptive strategies, demonstrating how dynamic parameter modulation prevents common generative failure modes.

\vspace{1mm}\noindent \textbf{Image-Conditioned (Fig.~\ref{fig:adaptive_noise_level}):} We observe that small or indistinct subjects (Panel a) are highly sensitive to noise; a high $\ell$ (row 3) often causes the object to vanish or distort. Our strategy correctly assigns a low $\ell$ (green boxes) to preserve semantic fidelity. Conversely, salient subjects (Panel b) remain robust under high noise, allowing us to safely introduce significant background and pose variations.
    
\vspace{1mm}\noindent \textbf{Text-Conditioned (Fig.~\ref{fig:adaptive_guidance_scale}):} Detailed captions (Panel a) already constrain the generation space; adding high guidance results in rigid, repetitive outputs. Our method relaxes guidance ($g$) to foster diversity. For simple captions (Panel b), high guidance is essential to prevent semantic drift, ensuring the generated image remains faithful to limited textual cues.
    
\vspace{1mm}\noindent \textbf{Image-Text-Conditioned (Fig.~\ref{fig:adaptive_ITC}):} This visualizes the synergy of both strategies. The grid demonstrates that our method consistently selects the optimal parameter combination (green/highligth!ed boxes). For instance, for semantically rich captions (Panel a \& b), a lower guidance scale encourages diverse generation, while simpler captions (Panel c \& d) benefit from a higher guidance scale to maintain semantic alignment. Meanwhile, low-saliency images (a \& c) require a lower noise level to preserve fine details, whereas high-saliency images (b \& d) tolerate a higher noise level to enable greater visual variation.

\subsection{Visualization of Feature Distribution}
\label{exp:qua_results}
\begin{figure*}
 \centering
 \includegraphics[width=0.96\linewidth]{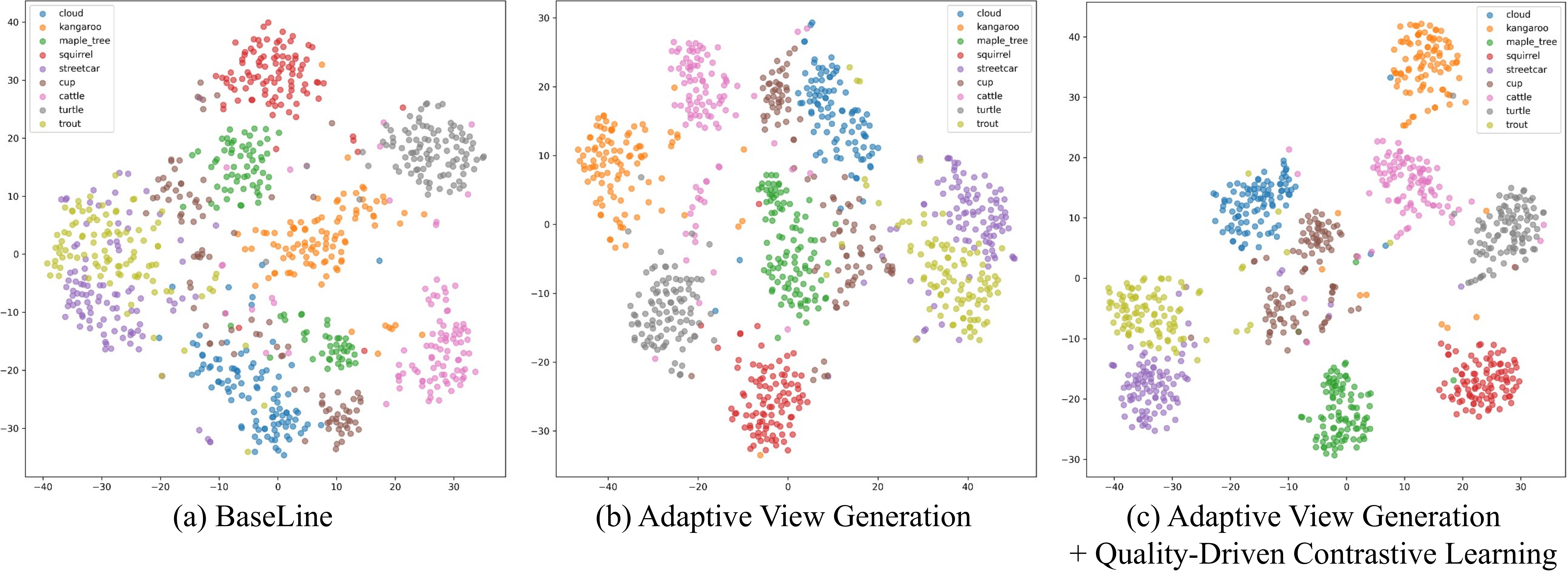}
 \caption{\textbf{t-SNE visualization of image features from 9 randomly selected CIFAR-100 classes}. (a) Baseline with standard augmentations shows scattered clusters. (b) Incorporating multi-source adaptive view generation improves separation. (c) Full GenView++ with quality-driven loss yields the most compact and distinct clusters.}
 \label{fig:tsne}
\vspace{-4mm}
\end{figure*}

To provide qualitative insigth! into the learned manifold, Fig.~\ref{fig:tsne} presents t-SNE visualizations of the embedding space for 9 randomly selected CIFAR-100 classes.
\textbf{(a) Baseline:} Standard augmentations result in diffuse clusters with significant overlap, indicating weak \textit{inter-class separability}.
\textbf{(b) +Adaptive Views:} Incorporating multi-source adaptive generation produces more distinct clusters. This confirms that diverse yet semantically consistent views help the model better capture class-specific semantics.
\textbf{(c) Full GenView++:} The addition of the quality-driven loss further compresses the clusters, significantly improving \textit{intra-class compactness} and sharpening decision boundaries. This progression visually validates that our quality-aware supervision effectively filters out noisy signals that would otherwise diffuse the feature distribution.

\subsection{Computational Cost and Efficiency}
\label{supp:discussion_cost}

We analyze the computational trade-off of GenView++, framing the generative cost as a one-time investment that yields long-term training dividends.

\vspace{1mm}\noindent \textbf{One-Time Offline Investment vs. Recurring Training Cost.}
A critical distinction must be made between \textit{setup cost} and \textit{training overhead}. Generating synthetic views indeed incurs GPU hours; however, this is a strictly one-off, offline investment. Unlike online augmentations (e.g., Mixup, AutoAugment) that consume CPU/GPU resources during every training iteration, our adaptive generation creates a static, reusable dataset. This dataset possesses high utility:
\textbf{First}, it can be reused indefinitely across diverse model architectures (ResNet, ViT), hyperparameter sweeps, and downstream tasks without regeneration.
\textbf{Second}, the initial generation cost is effectively amortized to zero when shared across the community or used for multiple training runs, making the marginal cost of adoption negligible.

\vspace{1mm}\noindent \textbf{Data Efficiency as a Performance Multiplier.}
The core value proposition of GenView++ is replacing data "quantity" with "quality". As evidenced in Table~\ref{tab:exp_naive}, GenView++ achieves superior accuracy (\textbf{65.62\%}) using only 0.15M generated views, outperforming an expansion with 0.3M real images from ImageNet-21K (64.10\%).
This result implies a higher information density per sample. Consequently, GenView++ enables models to converge to better representations with fewer training samples, potentially reducing the total training epochs required to reach target performance. This efficiency gain further offsets the initial generation cost.

\vspace{1mm}\noindent \textbf{Scalability.}
The generation process is embarrassingly parallel and inference-only (no backpropagation). It can be efficiently scaled on lower-end GPU clusters or inference-optimized hardware, ensuring that the barrier to adoption remains low even for resource-constrained environments.
\section{Conclusion}
In this paper, we propose \textbf{GenView++}, a unified framework that fundamentally addresses the data quality bottleneck in contrastive learning by harmonizing generative synthesis with discriminative supervision. 
To address the limitations of traditional augmentation, our Multi-Source Adaptive View Generation module dynamically modulates latent conditioning parameters based on input intrinsic properties, synthesizing diverse views that maintain strict semantic fidelity. Synergistically, our Quality-Driven Contrastive Learning mechanism functions explicitly reweight training pairs based on cross-modal alignment and visual novelty to filter out generative noise.
Crucially, we demonstrated that the generative overhead is a one-time offline investment, yielding a static, reusable pool of high-density data that significantly boosts sample efficiency without incurring online training costs.
Extensive experiments across vision and vision-language benchmarks confirm that GenView++ sets a new state-of-the-art. By effectively bridging controllable generation with quality-aware supervision, this work paves the way for more robust, data-centric representation learning in the era of foundation models.

{
    \small
    \bibliographystyle{ieeenat_fullname}
    \bibliography{main}
}

\setcounter{page}{1}
\setcounter{section}{0}
\setcounter{table}{0}
\setcounter{figure}{0}
\renewcommand{\thetable}{\Roman{table}}
\renewcommand{\thesection}{\Roman{section}}

\section{Implementation Details}

\subsection{Foreground Proportion Estimation}
To estimate the foreground proportion in Sec.~4.2.2, we adopt the pretrained \href{https://huggingface.co/laion/CLIP-ViT-H-14-laion2B-s32B-b79K}{CLIP ViT-H/14} backbone, which also serves as the conditional image encoder $\gV$ in Stable UnCLIP v2-1. Given a $224\times224$ input, this backbone produces 256 tokens of dimension 1280. For PCA-based feature extraction, we randomly sample 10,000 images from the training set. The threshold $\alpha$ in Eq.~7 is chosen such that foreground tokens constitute roughly 40\% of the total, ensuring a clear separation between foreground and background in Fig.~2 in the main paper.

\subsection{Adaptive View Generation Details}
\textbf{For IN1K, CF10, CF100, and TinyIN}, we generate one additional view per training image using the image-conditioned adaptive view generation method. The number of denoising steps $T$ is set to 20 for computational efficiency, and the classifier-free guidance scale \citet{ho2022classifier} is fixed to 10 to preserve image fidelity. The diversity of synthesized views is modulated by applying different noise perturbations to the image embeddings. To align with the input resolution of each dataset, the generated images are downsampled from their native $768^2$ resolution to $512^2$ (IN1K), $32^2$ (CF10 and CF100), and $64^2$ (TinyIN). 

\textbf{For the Conceptual Captions 3M (CC3M)} dataset \citet{sharma2018conceptual}, we adopt a hybrid data strategy: a portion of the data is enhanced with our multi-source adaptive view generation. 
Each enhanced image-text pair is expanded into a positive set (Eq.~3) using the three adaptive generation modes introduced in Secs.~4.2.2, 4.2.3, and 4.2.4. 
All images in these positive sets, including both original and generated views, are further processed by a standard augmentation pipeline.  
In the \textit{comparison with multimodal methods} setting (Sec.~5.2), we train on the full CC3M dataset ($\sim$3M image-text pairs), where a 1M subset is enhanced by the adaptive generation strategy.  
For the \textit{ablations on multimodal components} (Sec.~6.2), we train on a 1M subset of CC3M, within which only a portion (e.g., 100k or 500k samples) is enhanced, while the rest remain unmodified.

\section{Experimental Details}
\label{supp:details}
\subsection{Details for Vision Pretraining Comparisons}
\label{supp:details_vision_pretraining}
\textbf{Pretraining Setup.}
We conduct experiments using both Convolutional (ResNet-18, ResNet-50 \citet{he2016deep}) and Transformer (ViT-S, ViT-B \citet{dosovitskiy2020image}) backbones. ResNet-50 is the default architecture unless otherwise specified, while MoCov3 experiments also utilize ViTs.
All models are pretrained on the ImageNet-1K (IN1K) dataset \citet{deng2009imagenet}. We strictly adhere to the original training configurations of each baseline method to ensuring fair comparison. Our adaptive generative augmentation is applied to the entire IN1K training set.
Table~\ref{tab:exp_hyperparam_vision} provides a detailed summary of the hyperparameters used for each framework (MoCov2, SwAV, SimSiam, BYOL, MoCov3). We reproduce MoCov3 baselines using the official implementation.

\begin{table}[t!]
 \centering
 \caption{Pretraining hyperparameters used for comparison with vision representation learning methods on IN1K.}
 \resizebox{1\linewidth}{!}{
 \begin{tabular}{lcccccc}
 \toprule
 & MoCov2 & SwAV & SimSiam & BYOL&MoCov3 &MoCov3 \\ \midrule
 Optimizer & SGD & LARS& SGD & LARS&LARS&AdamW\\
 Learning Rate & 0.03& 0.6& 0.05& 4.8&1.2/9.6/4.8&2.4e-3\\
 Weight Decay & 1e-4& 1e-4 & 1e-4 & 1e-6&1e-6&0.1\\
 Momentum & 0.9 & 0.9 & 0.9 & 0.9 &0.9 &-\\
 Cosine Decay & \checkmark & \checkmark & \checkmark & \checkmark &\checkmark &\checkmark \\
 Batch Size & 256& 256& 256& 4096&512/4096/4096&4096/4096\\
 Loss& $\mathcal{L}^\text{img-img}_\text{NCE}$& $\mathcal{L}^\text{img-img}_\text{KL}$& $\mathcal{L}^\text{img-img}_\text{COS}$& $\mathcal{L}^\text{img-img}_\text{COS}$&$\mathcal{L}^\text{img-img}_\text{NCE}$&$\mathcal{L}^\text{img-img}_\text{NCE}$\\
 Epochs & 200& 200& 200& 200&100/300&300\\
 Backbone & ResNet50& ResNet50& ResNet50& ResNet50 &ResNet50 &VIT-S/ViT-B\\
 Embedding Dim & 2048& 2048& 2048& 2048 &2048&384/768\\
 Projection Dim & 128& 128& 2048& 256&256&256\\
 \bottomrule
 \end{tabular}
 }
 \label{tab:exp_hyperparam_vision}
\end{table}

\textbf{Linear Classification Protocol.}
After pretraining, we train a linear classifier on top of the frozen encoder using IN1K. To ensure fairness, we adopt identical training hyperparameters across all baselines: 90 epochs, a batch size of 1,024, and the SGD optimizer with momentum 0.9 and no weight decay. A cosine learning rate schedule \citet{loshchilov2016sgdr} is applied, with an initial learning rate of 0.4 for ResNet-based models and 12.0 for ViT-based models.

\textbf{Transfer Learning on MS-COCO.}
To assess generalization to pixel-level tasks, we transfer the pretrained models to MS-COCO \citet{lin2014microsoft} for object detection and instance segmentation using the Mask R-CNN \citet{he2017mask} framework with ResNet-50-FPN. Models are pretrained on IN1K for 200 epochs and fine-tuned on COCO train2017, evaluated on val2017 using the Detectron2 $1\times$ schedule \citet{wu2019detectron2} (90k iterations, with learning rate decay at 60k and 80k).

\subsection{Details for Vision-Language Comparisons}
\label{supp:vison_language_pretraining}
\vspace{1mm}\noindent \textbf{Pretraining Setup.}
Pretraining is conducted on the CC3M dataset \citet{sharma2018conceptual}, with 1M data enhanced by our multi-source adaptive view generation mechanism. We primarily adopt the ViT-B/16 backbone, and all reported results are based on ViT-B/16 unless otherwise noted (TripletCLIP uses ViT-B/32). Unless specified, vision–language models are pretrained for 40 epochs on 8 GPUs with a batch size of 64 per GPU. The AdamW optimizer is used with $\beta_1=0.9$, $\beta_2=0.98$, a base learning rate of $2\times10^{-4}$, weight decay of 0.1, and a 5-epoch warmup schedule.

\vspace{1mm}\noindent \textbf{Evaluation Benchmarks.}
The evaluation is performed on zero-shot image-text retrieval (MS COCO \citet{lin2014microsoft}, Flickr30k \citet{young2014image}, and Flickr8k \citet{hodosh2013framing}) and zero-shot/linear classification across ten diverse image classification datasets: ImageNet \citet{deng2009imagenet}, CIFAR-10 \citet{krizhevsky2009learning}, CIFAR-100 \citet{krizhevsky2009learning}, Aircraft \citet{maji2013fine}, DTD \citet{cimpoi2014describing}, Flowers \citet{nilsback2008automated}, Pets \citet{parkhi2012cats}, SUN397 \citet{xiao2010sun}, Caltech-101 \citet{fei2004learning} and Food-101 \citet{bossard2014food}.

\vspace{1mm}\noindent \textbf{Linear Classification Protocol.}
For downstream image classification, we pretrain the vision encoder on CC3M, freeze its weights, and train a linear classifier on top. The classifier is trained for 90 epochs using SGD with a momentum of 0.9. Following standard practice, the base learning rate is selected from a sweep over {0.001, 0.002, 0.005, 0.01, 0.02, 0.05, 0.1, 0.2, 0.3, 0.5}. Additional hyperparameters for the ten classification datasets used in the linear probe are summarized in Table~\ref{tab:exp_hyperparam_visionlanguage}.

\begin{table}[t!]
 \centering
 \caption{Hyperparameters for linear classification.}
 \resizebox{1\linewidth}{!}{
 \begin{tabular}{lcccccccccc}
 \toprule
 & IN1K & CF10 & CF100 & Aircraft & DTD & Flowers & Pets & SUN397 & Caltech-101 & Food-101 \\ \midrule
 GPUs & 8 & 2 & 2 & 1	 & 1 & 1	 & 1 & 1 &	1 & 2 \\
 Batch/GPU & 256 & 256 & 256 & 128 & 64 & 64 & 128 & 256 &	128 & 256 \\
 Drop last & True & True & True & False &False& False &False & True & False & True \\
 Classes & 1000 & 10 & 100 & 1000 & 47 & 102 & 37 & 397 & 101 & 101 \\
 \bottomrule
 \end{tabular}
 }
 \label{tab:exp_hyperparam_visionlanguage}
\end{table}

\subsection{Details for GenView++(V) Component Ablations}
\label{supp_ablation_v}
We detail the experimental setup for the vision-only ablation studies presented in Sec.~6.1 of the main text.

\textbf{Pretraining Setup.}
We use ImageNet-100 (IN100) \citet{tian2019contrastive}, a class-balanced subset of ImageNet-1K, as the pretraining dataset. The backbone is a ResNet-18 trained using the MoCo v3 framework.
We train for 100 epochs with a batch size of 512. We employ the LARS optimizer with a base learning rate of 1.2, weight decay of $1 \times 10^{-6}$, and momentum of 0.9. The learning rate follows a cosine decay schedule.
To construct the pool of synthetic positive views, we randomly sample 50,000 images from IN100 (maintaining class balance) to serve as condition inputs for our Image-Conditioned generation module.

\textbf{Evaluation Setup.}
We perform linear probing on the frozen encoder using the full IN100 dataset. The linear classifier is trained for 50 epochs. All other optimization settings (optimizer, augmentations) are consistent with the main ImageNet-1K linear evaluation protocol described in Sec.~\ref{supp:details_vision_pretraining}.

\begin{table}[t!]
 \centering
 \caption{Rule-based scoring instruction prompt $R$ for evaluating visual complexity of caption inputs.}
  \resizebox{\linewidth}{!}{
 \begin{tabular}{|p{8cm}|}
 \hline
 \{"role": "system", \\
 "content": "You are tasked with evaluating the visual complexity of a text prompt intended for image generation. Your goal is to assign a score from 1 to 4 that reflects how richly the prompt describes concrete visual elements. The more detailed and diverse the visual information, the higher the score should be.\\
 Visual complexity is determined by identifying constraints that directly affect how an image would be generated. These include the specificity of objects mentioned (such as type, quantity, or material), descriptive visual features (such as color, texture, or lighting), spatial relationships (such as layout or perspective), dynamic elements (such as actions or interactions), and stylistic cues (such as artistic genres or cultural elements). A prompt that contains a wide range of such features is considered more complex than one that simply names general categories.\\
 A prompt that only refers to broad object types without any additional visual information should be scored as 1. If it includes a small number of visual constraints—typically one to three—it should be scored as 2. A score of 3 is appropriate when the prompt contains a moderate level of detail, roughly four to six constraints. A prompt that includes more than six distinct and specific visual elements—such as combinations of materials, textures, color, spatial arrangement, and style—should be scored as 4. \\
 When evaluating, do not include abstract or emotional language that does not translate directly into visual features. Focus only on concrete and visualizable information. \\
 After analyzing the prompt, return a score from 1 to 4. Return the result in the following format: [score: ?]."\} \\ \hline

 \{"role": "user", \\
 "content": f"Now output the score of visual constraints for the following text prompt:\{ \}."\} \\ \hline
 
 \end{tabular}
 }
 \label{supp:rule}
 \vspace{-4mm}
\end{table}

\subsection{Details for GenView++(VL) Component Ablations}
\label{supp_ablation_vl}
We detail the setup for the vision-language ablation studies presented in Sec.~6.2.

\vspace{1mm}\noindent \textbf{Pretraining Setup.}
All models utilize a ViT-B/16 backbone initialized with CLIP weights.
Training is conducted on a 1M subset of CC3M. To evaluate the impact of our generation module, we replace a portion of the Models are trained for 25 epochs with a 2-epoch linear warmup. We use the AdamW optimizer with a peak learning rate of $4 \times 10^{-4}$ (base learning rate $2 \times 10^{-4}$), weight decay of 0.1, and $\beta$ coefficients $(\beta_1, \beta_2) = (0.9, 0.98)$. The training runs on 4 GPUs with a batch size of 64 per GPU.

\vspace{1mm}\noindent \textbf{Evaluation Setup.}
We evaluate the learned representations via linear probing on CIFAR-10 and ImageNet-100.
A linear classifier is trained on top of the frozen backbone for 90 epochs using SGD (momentum 0.9, no weight decay) with a batch size of 256.
Following StableRep \citet{tian2023stablerep}, we determine the optimal base learning rate for each dataset by sweeping over the range {0.001, 0.002, 0.005, 0.01, 0.02, 0.05, 0.1, 0.2, 0.3, 0.5}.

\subsection{Evaluation of Visual Complexity Estimation Precision}
\label{supp:caption_complexity}

To validate the reliability of our adaptive guidance strategy ($g_i^\text{ada}$), we assess the precision of the automated complexity estimator (DeepSeek-2-Chat-Lite). We constructed a reference benchmark using 500 randomly sampled captions from the training set, with ground-truth complexity scores generated by GPT-4o and verified by human annotators. Our automated scorer was then evaluated against this ground truth using the rubric in Table~\ref{supp:rule}. Results show 93.8\% accuracy (within a ±1 tolerance), confirming that our LLM-based scorer closely aligns with human perception and enables complexity-aware generation control.

\end{document}